\documentclass[10pt,journal,compsoc]{IEEEtran}

\ifCLASSOPTIONcompsoc
  \usepackage[nocompress]{cite}
\else
  \usepackage{cite}
\fi
\usepackage{booktabs}       %
\usepackage{multirow}       %
\usepackage{graphicx}       %
\usepackage{amsmath}        %
\usepackage{amssymb}        %
\usepackage{amsthm}         %
\usepackage{xcolor}         %
\usepackage{colortbl}       %
\usepackage{array}
\usepackage{tabularx}
\usepackage{enumitem}       %
\usepackage{microtype}      %
\usepackage{tikz}           %
\usetikzlibrary{positioning, arrows.meta, calc, fit, backgrounds, shapes.geometric}
\usepackage[caption=false,font=normalsize,labelfont=sf,textfont=sf]{subfig}
\usepackage{capt-of}
\usepackage{pifont}          %
\usepackage[most]{tcolorbox}   %
\usepackage{afterpage}      %
\usepackage{float}          %
\usepackage{needspace}
\usepackage[hidelinks]{hyperref}
\hypersetup{
  pdftitle={Cognition-Oriented Emotion Tracing from Causes to Consequences in Real-World Social Scenes},
  pdfauthor={Hao Li, Jinye Zhang, Bobo Li, Mong-Li Lee, Wynne Hsu, Zheng Wang, Hao Fei, and Min Zhang}}

\definecolor{theoryblue}{HTML}{2B4C7E}
\definecolor{theorybluebg}{HTML}{F8FAFC}
\definecolor{theorygold}{HTML}{B88720}
\definecolor{theoryteal}{HTML}{1E7B72}
\definecolor{theorytealbg}{HTML}{EEF8F6}
\definecolor{theorywine}{HTML}{8A5A63}
\definecolor{theorywinebg}{HTML}{FBF1F3}
\definecolor{failureink}{HTML}{8B2E2E}
\definecolor{failurebg}{HTML}{FDF5F5}
\definecolor{failuretagbg}{HTML}{F4E1E1}
\definecolor{conditionblue}{RGB}{69,112,168}
\definecolor{stateorange}{RGB}{211,105,31}
\definecolor{manifestgreen}{RGB}{45,136,86}
\definecolor{appviolet}{RGB}{117,74,164}
\definecolor{effectblue}{RGB}{91,121,164}
\definecolor{softgray}{RGB}{105,115,125}
\tcbset{
  theoryboxbase/.style 2 args={
    enhanced,
    breakable,
    colback=#2,
    colframe=#1!30!white,
    colbacktitle=#1!10!white,
    coltitle=#1!85!black,
    boxrule=0.5pt,
    drop fuzzy shadow=black!5!white,
    borderline west={2.5pt}{0pt}{#1},
    arc=2.5pt,
    left=6pt, right=6pt, top=5pt, bottom=4pt,
    before skip=8pt,
    after skip=9pt,
    fonttitle=\bfseries\small,
    attach boxed title to top left={yshift=-\tcboxedtitleheight/2, xshift=7pt},
    boxed title style={
      colback=#1!10!white,
      colframe=#1!30!white,
      boxrule=0.5pt,
      arc=1.5pt,
      left=6pt, right=6pt, top=2pt, bottom=2pt,
    },
  },
  appraisalbox/.style={theoryboxbase={theoryblue}{theorybluebg}},
  easibox/.style={theoryboxbase={theoryblue}{theorybluebg}},
  regulationbox/.style={theoryboxbase={theoryblue}{theorybluebg}},
  failurebox/.style={theoryboxbase={theoryblue}{theorybluebg}}
}

\newtcolorbox{defbox}[1][]{%
  enhanced,
  colback=theorybluebg,
  colframe=theoryblue!40!white,
  borderline west={2.5pt}{0pt}{theoryblue},
  boxrule=0.5pt,
  arc=2pt,
  left=7pt, right=7pt, top=6pt, bottom=6pt,
  before skip=7pt, after skip=7pt,
  fonttitle=\bfseries\small\sffamily,
  coltitle=theoryblue!90!black,
  attach boxed title to top left={yshift=-\tcboxedtitleheight/2, xshift=6pt},
  boxed title style={
    colback=white,
    colframe=theoryblue!40!white,
    boxrule=0.5pt,
    arc=1.5pt,
    left=5pt, right=5pt, top=1.5pt, bottom=1.5pt,
  },
  title={#1}
}

\newcommand{\TRACEScore}{\text{TRACE-Score}}
\newcommand{\traceframework}{\textsf{TRACE}}
\newcommand{\blueprint}{\textsf{Affective Blueprint}}
\newcommand{\benchmark}{\textsf{TRACE-Bench}}
\newcommand{\method}{\textsf{TRACER}}
\newcommand{\Observe}{\operatorname{Observe}}
\newcommand{\Appraise}{\operatorname{Appraise}}

\newtcbox{\obsref}{
  on line,
  enhanced,
  nobeforeafter,
  tcbox raise base,
  colback=black!3,
  colframe=black!28,
  coltext=black,
  boxrule=0.4pt,
  arc=1pt,
  boxsep=0pt,
  left=1.5pt,
  right=1.5pt,
  top=0.4pt,
  bottom=0.4pt
}
\newtcbox{\appref}{
  on line,
  enhanced,
  nobeforeafter,
  tcbox raise base,
  colback=black!3,
  colframe=black!28,
  coltext=black,
  boxrule=0.4pt,
  arc=1pt,
  boxsep=0pt,
  left=1.5pt,
  right=1.5pt,
  top=0.4pt,
  bottom=0.4pt
}
\newtcbox{\conclusionref}{
  on line,
  enhanced,
  nobeforeafter,
  tcbox raise base,
  colback=black!3,
  colframe=black!28,
  coltext=black,
  boxrule=0.4pt,
  arc=1pt,
  boxsep=0pt,
  left=1.5pt,
  right=1.5pt,
  top=0.4pt,
  bottom=0.4pt
}

\newcommand{\TraceTotalQAs}{3{,}746}
\newcommand{\TraceTotalVideos}{646}

\newcommand{\TraceEffectPhysicalCount}{452}

\newcommand{\TraceEffectMentalCount}{286}

\newcommand{\taskname}[2]{%
  \tikz[baseline=(tasknum.base)]
    \node[circle,fill=theoryblue,text=white,draw=none,
          inner sep=0pt,minimum size=1.15em,
          font=\sffamily\bfseries\scriptsize] (tasknum) {#1};%
  \nobreakspace\textrm{#2}%
}

\newcommand{\Description}[1]{}

\definecolor{accent}{HTML}{00A896}      %
\definecolor{accentdark}{HTML}{00796B}  %
\definecolor{accentlight}{HTML}{D9F3EF}  %

\newcommand{\runinhead}[1]{\par\addvspace{6pt}\noindent\textbf{#1}}

\newcommand{\field}[1]{\ifmmode\text{\texttt{#1}}\else\texttt{#1}\fi}

\newcommand{\tasktitle}[1]{\runinhead{#1.}\enskip}
\newcommand{\taskformulation}[2]{%
  \begingroup\small
  \begin{equation}
  \label{#1}
  #2
  \end{equation}
  \endgroup
  \nopagebreak[3]}

\newcommand{\datatitle}[1]{%
  \runinhead{#1.}\enskip}
\newcommand{\datastep}[2]{%
  \runinhead{Step #1: #2.}\enskip}
\newcommand{\pointpara}{\par\addvspace{6pt}}
\newcommand{\stagetitle}[2]{\runinhead{Stage #1: #2.}\enskip}

\newcommand{\movetitle}[1]{\runinhead{#1.}\enskip}

\newcommand{\Cext}{C_{\mathrm{ext}}}
\newcommand{\Cint}{C_{\mathrm{int}}}
\newcommand{\Ccause}{C_{\mathrm{cau}}}

\newcommand{\Astate}{A_{\mathrm{sta}}}
\newcommand{\Amanifest}{A_{\mathrm{man}}}
\newcommand{\Areg}{A_{\mathrm{reg}}}
\newcommand{\Emental}{E_{\mathrm{men}}}
\newcommand{\Eaffect}{E_{\mathrm{aff}}}
\newcommand{\Ephysical}{E_{\mathrm{phy}}}
\definecolor{resultink}{HTML}{B00020}
\newcommand{\expnum}[1]{#1}
\newcommand{\exphead}[1]{\runinhead{#1.}\enskip}

\begin{document}

\title{Cognition-Oriented Emotion Tracing from Causes to Consequences in Real-World Social Scenes}

\author{Hao~Li, Jinye~Zhang, Bobo~Li, Mong-Li~Lee, Wynne~Hsu, Zheng~Wang,~\IEEEmembership{Senior~Member,~IEEE},\\
Hao~Fei\textsuperscript{*},~\IEEEmembership{Senior~Member,~IEEE},
and Min~Zhang,~\IEEEmembership{Senior~Member,~IEEE}%
\IEEEcompsocitemizethanks{\IEEEcompsocthanksitem
Hao Li, Jinye Zhang, and Zheng Wang are with the School of Computer Science,
National Engineering Research Center for Multimedia Software,
Wuhan University, Wuhan 430072, China.
\IEEEcompsocthanksitem
Bobo Li is with the National University of Singapore, Singapore.
\IEEEcompsocthanksitem
Mong-Li Lee and Wynne Hsu are with the School of Computing,
National University of Singapore, Singapore.
\IEEEcompsocthanksitem
Hao Fei is with the Department of Computer Science, University of Oxford,
Wolfson Building, Parks Road, Oxford OX1 3QD, U.K.
\IEEEcompsocthanksitem
Min Zhang is with the Institute of Computing and Intelligence, Harbin
Institute of Technology (Shenzhen), Shenzhen, China.
\textsuperscript{*}Corresponding author: Hao Fei (haofei7419@gmail.com).}}

\markboth{Li \MakeLowercase{\textit{et al.}}: Cognition-Oriented Emotion Tracing}%
{Li \MakeLowercase{\textit{et al.}}: Cognition-Oriented Emotion Tracing}

\IEEEpubid{\parbox[t]{\textwidth}{\footnotesize\raggedright
This work has been submitted to the IEEE for possible publication.
Copyright may be transferred without notice, after which this version
may no longer be accessible.}}

\IEEEtitleabstractindextext{%
\begin{abstract}
Affective computing has progressed from traditional categorical emotion recognition with small specialized models to open-ended affective analysis with large models, lifting recognition performance to unprecedented levels. However, theories in affective science describe emotion as an unfolding process involving causally interacting components, with appraisal, regulation, and social interpretation shaping its emergence, expression, and consequences. Computational modeling and evaluation of this full process remain largely unexplored in the field. To this end, this paper proposes the TRACE framework, which formalizes an affective episode through three interrelated stages, Condition, Affect, and Effect, incorporating both observable cues and cognitive factors such as the subject's internal stance toward the situation and strategic regulation of emotional display. Based on this formalization, we present \textbf{\benchmark{}}, a benchmark for evaluating multimodal models on affective understanding in real-world social scenes, which comprises five tasks with widening structural spans, from grounded affect recognition, regulation decoding, and cause reasoning to effect reasoning and full-chain reconstruction, covering \TraceTotalQAs{} structured question-answer pairs over \TraceTotalVideos{} videos. A matched human--model comparison reveals a substantial gap to human performance. Across the full benchmark, the evaluated affect-specialized models also generally lag behind general-purpose MLLMs. Analysis of model outputs identifies recurring failure patterns, including treating displayed behavior as genuine feeling and fabricating plausible but unsupported events during long-chain generation. Motivated by these findings, we propose \textbf{\method{}}, a cognition-grounded structured reasoning method. Each derivation step couples a natural-language inference with explicit premise citations drawn from factual observations, cognitive appraisals, and established upstream conclusions. The linked steps form a traceable graph of intermediate and target conclusions. Experiments show that \method{} outperforms all evaluated model baselines on each of the five tasks.
Project page: \url{https://cogaffc.github.io/TRACE}.
\end{abstract}
}

\maketitle
\IEEEdisplaynontitleabstractindextext
\IEEEpeerreviewmaketitle

\ifCLASSOPTIONcompsoc
\IEEEraisesectionheading{\section{Introduction}\label{sec:intro}}
\else
\section{Introduction}
\label{sec:intro}
\fi

\begin{figure*}[t]
  \centering
  \includegraphics[width=\textwidth]{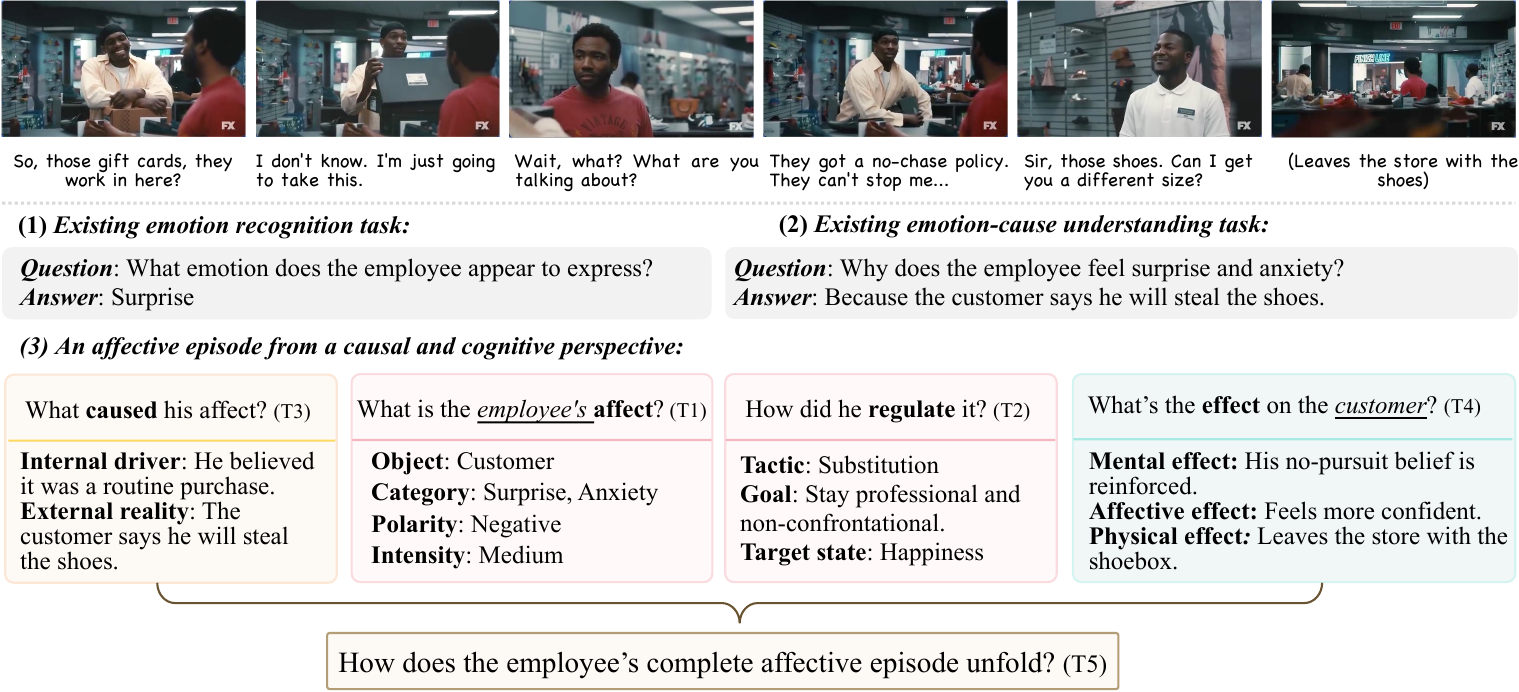}
  \caption{Existing affective benchmarks often evaluate local targets such as affective states or causes.
    \benchmark{} evaluates the full cognitive-affective chain: from the
    conditions that give rise to an emotion, through the regulation that
    reshapes its expression, to the consequences it produces for the subject or other participants.}
  \label{fig:teaser}
\end{figure*}
\looseness=-1

\IEEEPARstart{A}{ffective} computing has progressively expanded its analytical scope, evolving from the isolated recognition of surface signals to the contextual interpretation of human affect~\cite{zeng2009affectRecognitionSurvey}. Early paradigms formalized emotion as a closed classification problem, relying on small specialized models to predict categorical states or continuous dimensions from facial, acoustic, and linguistic cues~\cite{busso2008iemocap,kollias2023abaw,zadeh2016cmu}. As the field advanced, this scope broadened to incorporate conversational context~\cite{poria2019meld,majumder2019dialoguernn} and explicit causal triggers~\cite{xia2019ecpe,poria2021reccon}. More recently, the advent of multimodal large language models has propelled affective analysis beyond rigid labels, enabling open-ended emotion descriptions~\cite{lian2024emer,lian2025affectgpt}, causal explanations~\cite{wang2024ecem,lin2025eibench}, and generalized affective reasoning~\cite{cheng2024emotionllama,hu2025mtmeur,jiang2025mmeemotion}, lifting recognition performance to unprecedented levels. Nevertheless, despite this generative shift, the prevailing computational paradigm remains fundamentally observational: it maps perceptible signals to affective outcomes or provides descriptive accounts of emotional events. As surface-level recognition approaches human performance~\cite{zhang2026affectiveLLMSurvey}, the ceiling of this paradigm becomes apparent. By treating affect as a direct readout of observable behavior, it leaves cognitive mechanisms such as subjective appraisal, internal drivers, and strategic regulation largely unexplored.

In cognitive psychology and affective science~\cite{lazarus1991emotion,scherer2001appraisal,scherer2009dynamic}, affect has been studied not as an isolated output signal but as a dynamic process with complex cognitive involvement. Appraisal theories offer a causal perspective on emotion elicitation: events elicit different emotional responses depending on how they relate to a person's beliefs, desires, expectations, and coping potential~\cite{moors2013appraisalTheories,scherer2019emotionProcess}. The resulting affect may then be strategically managed before it becomes publicly visible~\cite{gross1998emerging}, and its display may further change what other people think, feel, and do~\cite{vankleef2009social}. In other words, an affective episode can be understood as a chain shaped by cognition: \textit{antecedent} $\rightarrow$ \textit{emotion} $\rightarrow$ \textit{consequence}. Capturing this full chain, rather than a single link on it, is a precondition for the next generation of affective applications: an empathetic dialogue agent needs to know where an emotion comes from in order to respond appropriately, a social robot needs to anticipate what its expression will bring about, and a mental-health support system needs to distinguish what a person displays from what the person feels. Computational research has started to adopt this cognitive view: textual commonsense studies infer the intents, reactions, and likely consequences behind events~\cite{rashkin2018event2mind,sap2019atomic,ghosal2022cicero}, and emotional-intelligence and Theory-of-Mind oriented benchmarks bring mental-state inference into evaluation~\cite{sabour2024emobench,yao2025emobenchm,hitemotion,core}. Yet each of these efforts covers only one section of the chain and remains scattered across modalities and task formats: the antecedent of affect is often reduced to an external trigger, omitting the internal stance that makes it matter; the regulation between felt and displayed affect is rarely modeled; and the downstream consequences of affect on others lack systematic evaluation. The result is a set of fragmented computational treatments that lack a unified definition of affective episodes from a formal cognitive perspective.

To address this gap, we introduce \traceframework{}, short for \textit{Tracing Emotions from Causes to Consequences}, to study cognitive-affective processes in real-world social scenes with complex interpersonal dynamics. \traceframework{} asks how a subject-centered affective episode can be formally represented, systematically evaluated, and computationally reconstructed from multimodal social interactions. We begin by defining the affective episode itself.

\begin{itemize}[
  leftmargin=0pt,
  label={},
  labelwidth=0pt,
  labelsep=0pt,
  itemindent=0pt,
  itemsep=0.65\baselineskip,
  topsep=0.4\baselineskip,
  parsep=0pt,
  partopsep=0pt
]
\item \textbf{\blueprint{}.}
We introduce an \blueprint{} that represents an affective episode through three causally connected stages: \textit{Condition} $\rightarrow$ \textit{Affect} $\rightarrow$ \textit{Effect}.
Condition captures the antecedent of affect, including the external reality and the subject's internal driver that together make the situation emotionally significant.
Affect records the felt state and its observable manifestation, along with the regulatory process by which the subject may reshape what is displayed, for example suppressing sadness to maintain composure.
Effect records the downstream consequences, mental, affective, and physical, that the episode produces for the subject or other participants.
Each transition in this chain is shaped by cognition, whether through the subject's interpretation of the triggering event, strategic regulation of display, or the observer's inference from what is shown.
Section~\ref{sec:blueprint:grounding} presents the theoretical grounding,
and Section~\ref{sec:blueprint:blueprint} gives the full formal definition.

\item \textbf{\benchmark{}.}
Based on this representation, \benchmark{} formulates affective process understanding as five complementary, diagnostic tasks whose structural span widens along the chain, from a single stage, to cross-stage causal relations, to the complete episode. \taskname{1}{Grounded Affect Recognition} evaluates whether a predicted affective state is supported by the subject's concrete manifestations. \taskname{2}{Regulation Decoding} evaluates whether a model can recover the cognitive regulation that mediates between the subject's felt state and outward manifestation, including why and how the displayed affect is reshaped. \taskname{3}{Affective Cause Reasoning} and \taskname{4}{Affective Effect Reasoning} trace an observed affect backward to its external reality and internal driver, and forward to its mental, affective, and physical consequences. \taskname{5}{Full Chain Reconstruction} jointly recovers Condition, Affect, and Effect. It evaluates whether the reconstructed episode is supported by the observed evidence and remains cognitively and causally correct. The five tasks share the same Blueprint and annotation framework, allowing different reasoning capabilities to be evaluated from a unified process perspective. Under this task design, we carefully curate and annotate \TraceTotalQAs{} structured question-answer pairs from \TraceTotalVideos{} video clips, spanning diverse social scenes, interpersonal relations, and configurations of the \blueprint{}.

\item \textbf{\method{}.}
Evaluation on \benchmark{} exposes two recurring failure patterns that motivate our method.
The first is \textit{surface reading}: a model reads affect from observable behavior without considering what the situation means to the person involved.
In Fig.~\ref{fig:teaser}, the employee maintains a friendly manner when the customer announces the theft.
Taken in isolation, his manner suggests a pleasant exchange.
From his perspective, however, the announcement threatens his job, and maintaining his composure is a way of coping with that threat.
Such a cognitive interpretation is needed to understand what gives rise to his affective state, what he actually feels, and why he regulates its expression, yet current models mostly remain at the level of observable behavior.
The second is \textit{factual drift}: when reconstructing an affective chain, a model treats interpretations it forms during reasoning, even when incorrect, as established facts and builds further conclusions on them, introducing plausible events that are not grounded in facts observable in the video.
To address these failures, we propose \method{}, which structures reasoning around explicit factual observations and cognitive analysis of what those observations mean to the characters involved.
Specifically, we first establish what occurs in the video, then analyze its significance from each character's perspective along four cognitive appraisal dimensions: relevance, implication, coping potential, and normative significance.
With this factual and cognitive basis, we derive the requested constructs through a task-specific reasoning topology grounded in the Blueprint.
The topology specifies the dependencies supporting each inference, which the model makes explicit by citing the relevant observations, appraisal readings, or established conclusions as premises.
Evaluation on \benchmark{} shows that \method{} outperforms all evaluated model baselines on each of the five tasks. Diagnostic experiments show reduced factual fabrication under \method{} and improved State attribution with supplied appraisal records.
\end{itemize}

This work makes three contributions.
\begin{itemize}[leftmargin=*,nosep]
\item We introduce the Affective Blueprint to formalize affective episodes in complex social scenes as three interrelated stages, Condition, Affect, and Effect, with cognition shaping the relations among them.
\item We introduce \benchmark{}, a multimodal benchmark with \TraceTotalQAs{} structured QA pairs over \TraceTotalVideos{} videos and five tasks that evaluate affective chain understanding at widening structural spans, from single-stage analysis through cross-stage reasoning to full-chain reconstruction.
\item We propose \method{}, a Blueprint-guided structured reasoning method that grounds affective inference in factual observations and character-specific cognitive appraisals.
\end{itemize}

\section{Related Work}
\label{sec:related}

\subsection{Affective Recognition and Extraction}
\label{sec:related:perceptual}

Affective computing has long relied on benchmarks that make affect
measurable through localized targets.
Representative datasets such as IEMOCAP~\cite{busso2008iemocap},
MELD~\cite{poria2019meld}, and CMU-MOSI~\cite{zadeh2016cmu} standardize
categorical or sentiment-oriented prediction from speech, facial
behavior, and language.
Dimensional settings further evaluate valence and arousal
signals~\cite{russell1980circumplex}, as in the ABAW
series~\cite{kollias2023abaw}.
{EMOTIC~\cite{kosti2019emotic} combines person and scene context to
recognize categorical and dimensional emotions in images.}
Recent multimodal resources extend the input space to richer and larger
settings, including face-audio integration~\cite{yang2025omniemotion}
and privacy-aware de-identified observation~\cite{xie2025deemo}.
{MERBench~\cite{lian2024merbench} provides a unified evaluation of feature
selection, multimodal fusion, and robustness for emotion recognition.}
These benchmarks are important because they establish whether a model
can recognize affective evidence, yet their evaluation target is still
centered on the current state, polarity, or expressed signal.
A related line moves from recognition to extraction, asking which part
of the context explains an affective state.
Emotion-Cause Pair Extraction~\cite{xia2019ecpe},
RECCON~\cite{poria2021reccon}, MECPE~\cite{wang2021mecpe}, and
SemEval-2024 Task~3~\cite{wang2024semeval_meca} recover the utterance,
event, or pair associated with an emotion in dialogue.
ECEM~\cite{wang2024ecem} adds free-form causal explanations, and
MTMEUR~\cite{hu2025mtmeur} expands the format to progressive question
answering over affective cues and triggering factors.
Together, recognition and extraction benchmarks make affective
evaluation concrete: a model predicts a state, identifies a cue, or
selects an explanatory fragment.
The resulting targets are useful and measurable, but they remain local
with respect to the affective episode.
They do not jointly evaluate the external situation, the subject's
internal driver, the relation between felt state and manifestation, and
the effects that follow in the surrounding~interaction.

\subsection{Generative Affective Understanding}
\label{sec:related:generative}

The rise of MLLMs has shifted affective evaluation from closed-form
prediction toward open-ended output.
Recent surveys of affective computing in the LLM era note that
heterogeneous affective understanding and generation tasks are
increasingly cast as instruction-following or sequence-generation
problems~\cite{zhang2026affectiveLLMSurvey}.
This matters for benchmark design because the expected output is no
longer limited to a class label; it may also be an explanation, caption,
judgment, or response.
EMER~\cite{lian2024emer} makes this shift explicit by asking models not
only to predict emotions, but also to provide the evidence and reasoning
behind the prediction, thereby using explanations to reduce label
ambiguity and support more fine-grained affective labels.
AffectGPT~\cite{lian2025affectgpt} further develops this direction with
descriptive emotion annotations, MER-Caption, and MER-UniBench, aligning
multimodal emotion understanding with the free-form output style of
MLLMs.
Emotion-LLaMA~\cite{cheng2024emotionllama} focuses on instruction-tuned
multimodal emotion recognition and reasoning with audio, visual, and
textual inputs, while EmoLLMs~\cite{cheng2024emollms} studies
instruction-following LLMs for broader affective analysis across
classification and regression tasks.
In parallel, emotional intelligence benchmarks such as
EmoBench-M~\cite{yao2025emobenchm} and
MME-Emotion~\cite{jiang2025mmeemotion} broaden evaluation from single
tasks to multi-scenario understanding and reasoning, including the causes
of an emotion. Their tasks are nevertheless scored independently, and none
represents how a felt state is regulated in display or what it brings about
in others. \benchmark{} instead organizes its tasks around one
subject-centered episode, so that Condition, State, Manifestation,
Regulation, and Effect are evaluated as components of one affective
episode rather than as separate abilities.

\subsection{Cognitive and Social Reasoning}
\label{sec:related:social}

A third line of work studies the social and cognitive reasoning needed
to interpret human behavior beyond surface signals.
Textual commonsense work, including Event2Mind~\cite{rashkin2018event2mind},
ATOMIC~\cite{sap2019atomic}, and CICERO~\cite{ghosal2022cicero},
studies intent, reaction, motivation, and likely subsequent events.
In video and multimodal settings, Social-IQ~\cite{zadeh2019socialiq}
evaluates question answering for artificial social intelligence,
MovieGraphs~\cite{vicol2018moviegraphs} represents movie clips with
graphs of characters, relationships, interactions, attributes, and
motivations, and DramaQA~\cite{choi2021dramaqa} uses
character-centered hierarchical QA to evaluate video story
understanding.
Multimodal intent benchmarks such as
MIntRec2.0~\cite{zhang2024mintrec20} further examine how language,
vision, and audio support intent recognition in multi-party
conversations.
Theory-of-mind benchmarks make the latent cognitive target more explicit:
MMToM-QA~\cite{jin2024mmtom} evaluates belief and goal inference in
household activities, MuMA-ToM~\cite{shi2025mumatom} extends the setting
to multi-agent interactions and beliefs about others' goals, and
MoMentS~\cite{villa2025moments} evaluates ToM abilities in realistic
narrative videos.
EmoBench~\cite{sabour2024emobench} and
ToMBench~\cite{chen2024tombench} likewise place affective or social
understanding within broader abilities such as emotional intelligence,
intent attribution, and mental-state inference.
These benchmarks clarify why affective computing in complex social
scenes cannot be reduced to recognizing visible expressions.
Their primary targets include intentions, beliefs, relations, story
states, and general social judgments.
\benchmark{} centers these capabilities on a subject-centered affective
episode, evaluating whether the triggering condition, the felt affective
state, its manifestation, possible regulation, and the resulting mental,
affective, and physical effects are mutually consistent under the
Affective Blueprint.

\section{Preliminaries}
\label{sec:blueprint}

This section draws on established theories from cognitive science and
social psychology to analyze the internal composition of an affective
process, and formalizes this analysis into a representation with
concrete stages and attributes. Section~\ref{sec:blueprint:grounding}
provides the theoretical grounding, and
Section~\ref{sec:blueprint:blueprint} translates it into the Affective
Blueprint.

\subsection{Theoretical Grounding}
\label{sec:blueprint:grounding}

\afterpage{%
\begin{figure}[!t]
\centering
\includegraphics[width=\columnwidth]{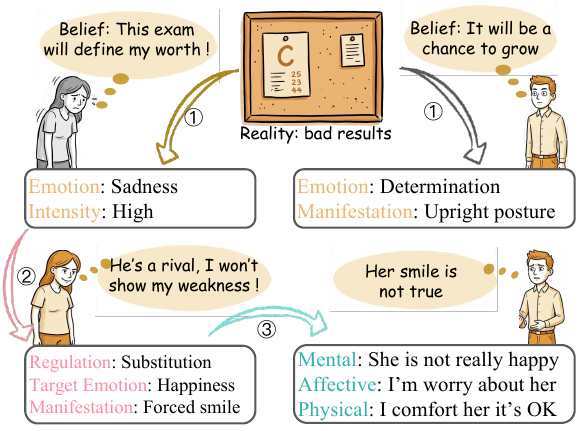}
\caption{Theoretical grounding of affect as a causally structured
process. The example illustrates how internal stance shapes affective
meaning, how regulation can separate felt affect from display, and how
displayed affect can influence others.}
\label{fig:theory_scenarios}
\end{figure}}

\pointpara
\noindent\textbf{\textsc{Theory~1:}~\textit{Cognitive Appraisal Theory}~\cite{lazarus1991emotion,scherer2001appraisal,ortony1988occ}}

\begin{tcolorbox}[appraisalbox]
\small
A situation acquires affective significance through appraisal of the
relation between the external situation and the subject's internal stance.
\end{tcolorbox}

Affective responses depend on both what happens and the psychological
state from which the subject encounters it. Appraisal theory explains
this dependency: the significance of an event emerges through its
evaluation in relation to the subject's current beliefs, goals, concerns,
and coping possibilities. The same external event can therefore produce
different affective responses when subjects bring different internal
states to it. As illustrated in Fig.~\ref{fig:theory_scenarios}, a poor exam
result may evoke sadness for a student who treats it as evidence of
personal failure, while eliciting determination when it is understood as
an opportunity for growth. \textit{Modeling an affective episode therefore
requires recovering both the external situation and the subject-specific
internal conditions.}

\pointpara
\noindent\textbf{\textsc{Theory~2:}~\textit{Emotion Regulation Theory}~\cite{gross1998emerging,gross2015emotion}}

\begin{tcolorbox}[regulationbox,unbreakable]
\small
Emotion regulation can modify affective experience and expression,
allowing outward manifestation to diverge from the felt state.
\end{tcolorbox}

Even after an emotion has emerged, its outward manifestation may differ
from what the subject feels. The mapping from felt affect to observable
expression can be shaped by emotion regulation. A subject may suppress,
amplify, substitute, or fabricate an affective display in response to
social goals, relationship concerns, or situational
expectations~\cite{ekman1969repertoire,hochschild1979emotion}.
The lower-left part of Fig.~\ref{fig:theory_scenarios} illustrates this
relation. Although the student remains sad, the presence of a rival
motivates her to conceal the sadness and display a forced smile. This
illustrates that an observable manifestation cannot always be interpreted
as a direct readout of the underlying affective state. \textit{Therefore,
understanding an affective process requires considering whether and how
emotion regulation shapes the transition from what is felt to what is
shown.}

\pointpara
\noindent\textbf{\textsc{Theory~3:}~\textit{Emotions as Social Information (EASI)}~\cite{vankleef2009social}}

\begin{tcolorbox}[easibox]
\small
Emotions provide social information that can influence what others
understand, feel, and do during interaction.
\end{tcolorbox}

Affective experiences extend beyond individual expression and can shape
the subsequent course of social interaction. The Emotions as Social
Information (EASI) framework explains that a subject's affective response
provides socially relevant information that can influence how others
understand the person and situation, how they feel in response, and how
they subsequently behave~\cite{vankleef2009social,vankleef2022social}.
Observers interpret available affective cues together with the situational
context, updating their beliefs and forming affective responses that can
further guide their behavior toward the
subject~\cite{demelo2014reading}. Affect can therefore
propagate across individuals, producing cognitive, affective, and
behavioral changes in others.

The lower-right part of Fig.~\ref{fig:theory_scenarios} illustrates this
interpersonal dynamic. In the example, another student forms the belief
that she is concealing sadness (mental effect), feels concern for her
(affective effect), and steps forward to offer comfort (physical effect).
\textit{Therefore, understanding an affective process requires tracing
how a subject's affective response gives rise to downstream mental,
affective, and physical consequences in others.}

\begin{figure}[!t]
\centering
\includegraphics[width=\columnwidth]{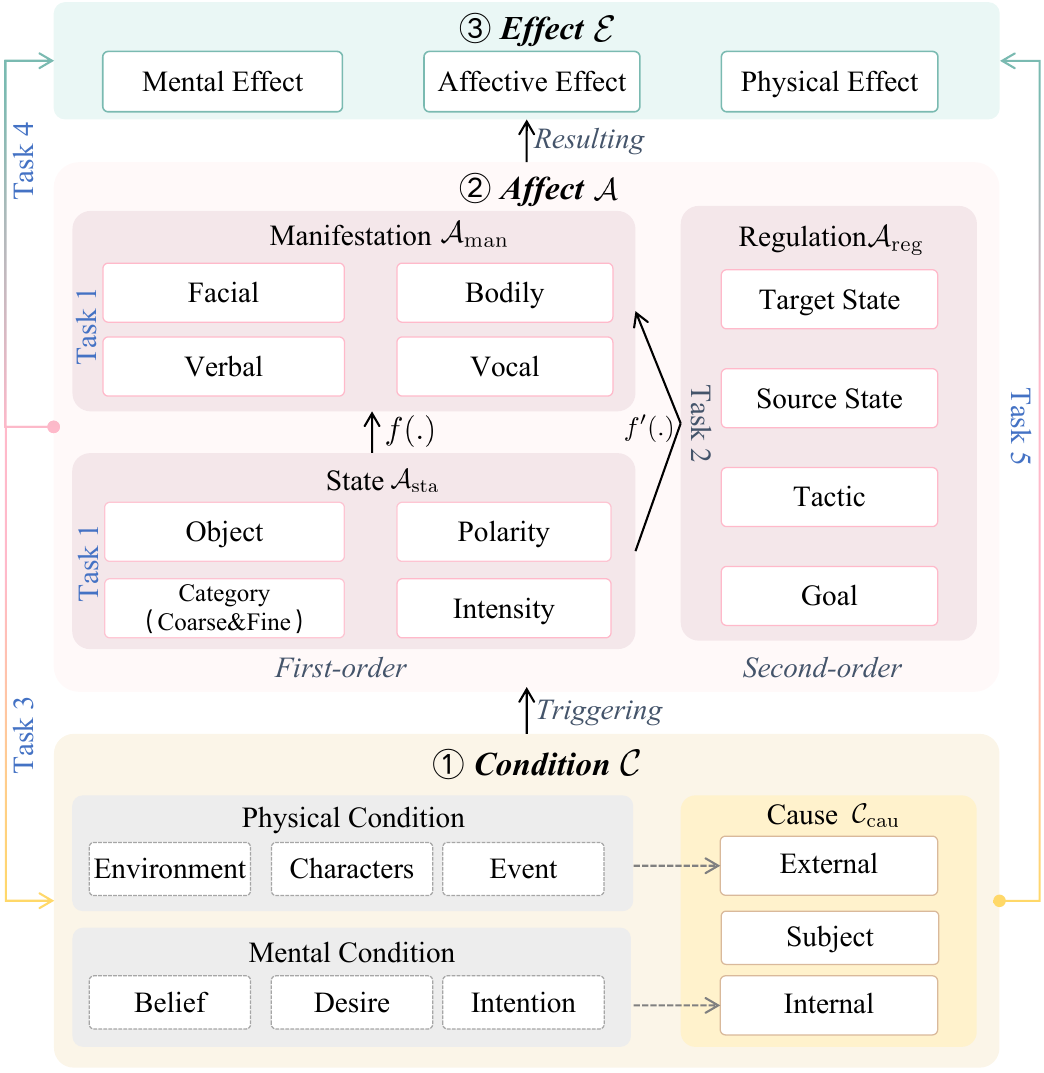}
\caption{The \textbf{Affective Blueprint}. A subject-centered
affective episode is represented through three causally connected stages:
\textit{Condition}, \textit{Affect}, and \textit{Effect}. Each stage is decomposed into several components, each represented by a set of fields.}
\label{fig:blueprint}
\end{figure}

Together, the three theories each anchor one span of the chain:
appraisal theory explains how affect is triggered (Condition),
regulation theory explains how a feeling reaches its public form
(Affect), and EASI explains how a display acts on others (Effect).
They jointly establish the cognitive and causal account of affect that
we formalize~below.

\subsection{The Affective Blueprint}
\label{sec:blueprint:blueprint}

Building on these theoretical foundations, we construct the Affective Blueprint
$\mathcal{B}$, our formal representation of a subject-centered affective
episode in real-world social interaction. The Blueprint includes three
aspects that our formulation treats as jointly necessary: the conditions
through which a situation acquires affective significance, the affective
response together with its regulation and manifestation, and the downstream
consequences that unfold in the interaction. We organize them into three
causally connected stages, Condition $\mathcal{C}$, Affect $\mathcal{A}$,
and Effect $\mathcal{E}$:
\begin{equation}
\label{eq:blueprint_episode}
\mathcal{B} = \langle \mathcal{C},\; \mathcal{A},\; \mathcal{E} \rangle.
\end{equation}
Fig.~\ref{fig:blueprint} shows the resulting structure.

\afterpage{\begin{table*}[!t]
\centering
\caption{Field-level schema of the Affective Blueprint. Each row specifies one
field and its admissible value space. Effect fields describe the impact of the
subject's affect. Effect recipients are identified within
the corresponding effect fields.}
\label{tab:blueprint_fields}
\begingroup
\footnotesize
\setlength{\tabcolsep}{4pt}
\renewcommand{\arraystretch}{1.14}
\newcommand{\bpstageband}[3]{%
  \rowcolor{#1!10}%
  \multicolumn{4}{@{}l@{}}{%
    \hspace{4pt}\textcolor{#1}{\textbf{\sffamily #2}}%
    \hspace{0.75em}\textcolor{black}{\textit{#3}}}\\[-0.2pt]}
\newcommand{\bpcomponent}[2]{%
  \textbf{#1}\,\textcolor{black}{($#2$)}}
\newcommand{\bpfieldname}[1]{\textcolor{black}{#1}}
\newcommand{\bpform}[1]{\textcolor{black}{#1}}
\newcommand{\bpdivider}{%
  \arrayrulecolor{black!28}\cmidrule(r){2-4}\arrayrulecolor{black}}
\begin{tabularx}{\textwidth}{@{}l@{\hspace{5pt}}l@{\hspace{6pt}}>{\raggedright\arraybackslash}X@{\hspace{6pt}}>{\raggedright\arraybackslash}p{140pt}@{}}
\toprule
\textbf{Component} & \textbf{Field} & \textbf{Definition} & \textbf{Value space} \\
\midrule

\bpstageband{theorygold}{I \enspace CONDITION}{What triggers an affect?}
\multirow{3}{*}{\bpcomponent{Cause}{\Ccause}}
  & \bpfieldname{subject}
  & Focal person in the affective episode
  & \bpform{Free-form text} \\
  & \bpfieldname{external\_reality}
  & Objective event triggering the subject's affect
  & \bpform{Free-form text} \\
  & \bpfieldname{internal\_driver}
  & Subject's belief, desire, or intention motivating affect
  & \bpform{Free-form text} \\
\addlinespace[2.8pt]

\bpstageband{theorywine}{II \enspace AFFECT}{What is the subject's affective state, and how is it expressed and regulated?}
\multirow{5}{*}{\bpcomponent{Affective State}{\Astate}}
  & \bpfieldname{object}
  & Person toward whom the affect is directed
  & \bpform{Free-form text} \\
  & \bpfieldname{coarse\_category}
  & Category of the subject's affective state
  & \bpform{12 categories} \\
  & \bpfieldname{fine\_description}
  & Contextual nuance or blend of affect
  & \bpform{Free-form text} \\
  & \bpfieldname{polarity}
  & Valence of the affective state
  & \bpform{\{Positive, Negative, Neutral, Mixed\}} \\
  & \bpfieldname{intensity}
  & Strength of the affective state
  & \bpform{\{High, Medium, Low, None\}} \\
\bpdivider
\multirow{4}{*}{\bpcomponent{Manifestation}{\Amanifest}}
  & \bpfieldname{facial\_expression}
  & Observable facial movement or configuration
  & \bpform{Free-form text} \\
  & \bpfieldname{body\_language}
  & Observable posture, gesture, movement, or spatial behavior
  & \bpform{Free-form text} \\
  & \bpfieldname{vocal\_cues}
  & Tone, pitch, volume, pace, or rhythm
  & \bpform{Free-form text} \\
  & \bpfieldname{verbal\_content}
  & Subject's words or speech act
  & \bpform{Free-form text} \\
\bpdivider
\multirow{4}{*}{\bpcomponent{Regulation}{\Areg}}
  & \bpfieldname{tactic\_type}
  & Regulatory relation between source and target
  & \bpform{See Eq.~\eqref{eq:tactics}} \\
  & \bpfieldname{goal}
  & Purpose of regulating the affective state
  & \bpform{Free-form text} \\
  & \bpfieldname{source\_affect\_state}
  & Affective state before regulation
  & \bpform{12 categories} \\
  & \bpfieldname{target\_affect\_state}
  & Affective state the subject intends to manifest
  & \bpform{12 categories} \\
\addlinespace[2.8pt]

\bpstageband{theoryteal}{III \enspace EFFECT}{How does the subject's affect influence those present?}
\bpcomponent{Mental Effect}{\Emental}
  & \bpfieldname{mental\_effect}
  & Impact on a recipient's mental state
  & \bpform{Free-form text} \\
\bpdivider
\bpcomponent{Affective Effect}{\Eaffect}
  & \bpfieldname{affective\_effect}
  & Impact on a recipient's affective state
  & \bpform{Free-form text} \\
\bpdivider
\bpcomponent{Physical Effect}{\Ephysical}
  & \bpfieldname{physical\_effect}
  & Impact on a recipient's action or speech
  & \bpform{Free-form text} \\
\bottomrule
\end{tabularx}

\endgroup
\end{table*}
}

\stagetitle{I}{Condition}
Given a \textbf{Subject} ($C_{\mathrm{sub}}$) whose affective process is
being analyzed, the Condition stage provides the broader context from
which the subject's affective response emerges. Physical Condition
describes the external situation through its environment, characters,
and events, while Mental Condition captures the subject's beliefs,
desires, and intentions~\cite{BDI}. For a specific
affective episode, we further identify the antecedents most directly
relevant to the response as its affective cause. Specifically,
\textbf{External Reality} ($C_{\mathrm{ext}}$) captures the external
circumstance implicated in the response, while \textbf{Internal Driver}
($C_{\mathrm{int}}$) captures the subject-specific internal stance that
gives this circumstance affective significance. We represent the
affective cause as
\begin{equation*}
C_{\mathrm{cau}} = (C_{\mathrm{ext}}, C_{\mathrm{sub}}, C_{\mathrm{int}}).
\end{equation*}

\stagetitle{II}{Affect}
The Affect stage comprises two first-order components, Affective State
and Manifestation, together with a second-order analysis of their relation,
Regulation.
\textbf{Affective State} ($A_{\mathrm{sta}}$) records what the subject feels
through four fields: object, category, polarity, and intensity. Object
specifies the person toward whom the affect is directed. Category identifies
the affect at coarse and fine-grained levels, while polarity and intensity
characterize its valence and strength.
\textbf{Manifestation} ($A_{\mathrm{man}}$) records the outward expressions
corresponding to $A_{\mathrm{sta}}$ across facial, bodily, verbal, and vocal
channels. Facial and bodily channels capture visible expressions and
movements, while verbal content captures what the subject says and vocal
cues characterize the utterance's acoustic and prosodic delivery.

\textbf{Regulation} ($A_{\mathrm{reg}}$) provides a second-order analysis
of the relation between $A_{\mathrm{sta}}$ and $A_{\mathrm{man}}$. It records
the regulation tactic, its goal, and the source and target affective states.
The source state $s$ corresponds to the subject's felt affective state
$A_{\mathrm{sta}}$, while the target state $t$ denotes the affective state
intended for outward display through $A_{\mathrm{man}}$. In
Fig.~\ref{fig:theory_scenarios}, for example, the student remains sad but
displays a forced smile. Sadness is the source state, the smile is the
manifestation, and happiness is the target state. Under Suppression, the
target retains the source category even when its outward expression becomes
barely perceptible. The tactic characterizes the relation between the source
and target states.

Let $s_c$ and $t_c$ denote their respective categories, and let $N$ denote
Neutral. For a shared non-neutral category,
$r_i\in\{\mathrm{lower},\mathrm{comparable},\mathrm{higher}\}$ describes the
intensity conveyed by the manifestation relative to the felt intensity.
The five tactics are defined as follows:
\par
\begingroup
\footnotesize\setlength{\arraycolsep}{2pt}
\begin{equation}
\label{eq:tactics}
\operatorname{tactic}(s,t,r_i) =
\begin{cases}
\text{None}, & s_c=t_c=N,\\
 & \text{or }s_c=t_c\neq N,\ r_i=\mathrm{comparable},\\[2pt]
\text{Suppression}, & s_c=t_c\neq N,\ r_i=\mathrm{lower},\\[2pt]
\text{Amplification}, & s_c=t_c\neq N,\ r_i=\mathrm{higher},\\[2pt]
\text{Substitution}, & s_c\neq N,\ t_c\notin\{s_c,N\},\\[2pt]
\text{Fabrication}, & s_c=N,\ t_c\neq N.
\end{cases}
\end{equation}
\endgroup
For a multi-label $A_{\mathrm{sta}}$, $s_c$ denotes the particular affective
category involved in the regulation. We record the target category and
encode its intensity relative to the source through the regulation tactic,
rather than assigning a separate target-intensity value.
The Affect stage is thus represented as
\begin{equation*}
\mathcal{A} = (A_{\mathrm{sta}}, A_{\mathrm{man}}, A_{\mathrm{reg}}).
\end{equation*}

\stagetitle{III}{Effect}
The Effect stage records the downstream consequences of the Subject's
affective response within the interaction. We distinguish three channels.
\textbf{Mental Effect} ($E_{\mathrm{men}}$) records what a recipient comes
to believe, desire, or intend. \textbf{Affective Effect} ($E_{\mathrm{aff}}$)
records the affect experienced by the recipient as a consequence.
\textbf{Physical Effect} ($E_{\mathrm{phy}}$) records the recipient's
resulting action or behavioral response. Each channel is associated with
its own recipient, so different effects within the same affective episode
may involve different individuals.
The recipient may also be the Subject. For example, a person who remains
angry with a colleague may later refuse the colleague's request for help
because of that anger. Here, the Physical Effect is the Subject's own
subsequent behavior. The Effect stage is represented as
\begin{equation*}
\mathcal{E} = (E_{\mathrm{men}}, E_{\mathrm{aff}}, E_{\mathrm{phy}}).
\end{equation*}

Table~\ref{tab:blueprint_fields} summarizes the fields and value forms used
to represent these components.

\section{\benchmark{}: Benchmarking Affective Chain Inference}
\label{sec:bench}

Building on the Affective Blueprint, we develop \benchmark{} to evaluate
affective chain inference in multimodal social videos. We first
formulate five tasks covering within-stage reasoning, cross-stage
reasoning, and full-chain reconstruction, then describe the data
construction process and evaluation protocol.

\subsection{Task Formulation}
\label{sec:task}

Each task takes a multimodal social video \(v\), including its visual
content, audio, and aligned subtitles, as the base input. A task instance
specifies three parts:
\begin{itemize}[leftmargin=1.4em,itemsep=1pt,topsep=2pt]
\item \textbf{Context} ($\mathrm{ctx}_t$) identifies the focal subject and
episode.
\item \textbf{Known Constructs} ($K_t$) are the Blueprint constructs whose
values are provided by the question.
\item \textbf{Queried Constructs} ($Q_t$) specify which Blueprint constructs
the model is asked to infer and, for a queried Effect, its channel and
recipient.
\end{itemize}
A known construct may be supplied through only a subset of its fields,
as specified for each task. The answer template specifies the requested
fields and any supporting evidence required by the task. The model
generates the answer as
\begin{equation}
\label{eq:unified_task}
\widehat{Y}_t=f\!\left(v,\mathrm{ctx}_t,K_t,Q_t\right).
\end{equation}
The five tasks cover within-stage reasoning (Tasks~1--2), cross-stage
reasoning (Tasks~3--4), and full-chain reconstruction (Task~5). The mappings
below summarize their given and queried constructs, with the shared video
input $v$ omitted. Construct symbols denote task-level roles, with the
supplied or predicted fields specified in the accompanying text and answer
template. Square brackets indicate annotation-dependent constructs.

\begin{subequations}
\label{eq:task_mappings}

\tasktitle{Task 1. Grounded Affect Recognition}
Given \(\mathrm{ctx}_1\), which specifies the focal subject and episode,
the model jointly predicts Affective State \(\Astate\) and
Manifestation \(\Amanifest\):
\taskformulation{eq:task1_mapping}{
(\mathrm{ctx}_1)
\longrightarrow
Q_1=\{\Astate,\Amanifest\}
}
No Blueprint constructs are supplied. Task~1 evaluates joint recognition
of the subject's felt state and observable expressions from multimodal
evidence.

\tasktitle{Task 2. Regulation Decoding}
Task~2 asks whether and how the observed manifestation is regulated
relative to the felt affective state. The task supplies the category under
regulation, its intensity, and the affective object as known information
about \(\Astate\). Within the episode specified by
\(\mathrm{ctx}_2\), the model infers regulation \(\Areg\):
\taskformulation{eq:task2_mapping}{
(\mathrm{ctx}_2;\,\Astate)
\longrightarrow
Q_2=\{\Areg\}
}
The model reads the display from the video and determines whether it has
been reshaped relative to the given state. Its answer identifies the tactic
and, when regulation occurs, the goal and target affective state intended
for outward display. In addition to the queried Regulation construct,
Task~2 requires \textbf{Regulation Evidence} ($M_{\mathrm{reg}}$), comprising
the manifestation cues in $\Amanifest$ that specifically support the
inferred regulation. $M_{\mathrm{reg}}$ is scored on regulated instances
and is not an additional Blueprint construct. The source state is fixed by
the supplied \(\Astate\). Whereas Task~1 jointly identifies what is felt
and shown, Task~2 examines the regulatory relation between them.

\tasktitle{Task 3. Affective Cause Reasoning}
Task~3 reasons backward from the subject's Affect to its cause. Given
\(\Astate\), \(\Amanifest\), and any available annotation of \(\Areg\), together
with \(\mathrm{ctx}_3\), the model infers the relevant antecedents in the
Condition stage:
\taskformulation{eq:task3_mapping}{
(\mathrm{ctx}_3;\,\Astate,\Amanifest,[\Areg])
\longrightarrow
Q_3=\{\Cext,[\Cint]\}
}
The supplied \(\Astate\) includes category, intensity, polarity, and
object, but excludes the fine-grained description. \(\Amanifest\)
includes the annotated channels. The model must ground the affective outcome
in its external antecedent
and, when the annotation supports it, infer the Internal Driver that
gives the event its affective significance. External Reality is
required in every Task~3 instance, while Internal Driver is included only
when supported by the annotation.

\tasktitle{Task 4. Affective Effect Reasoning}
Task~4 reasons forward from the subject's Affect to its Effect. Given
\(\Astate\), \(\Amanifest\), and any available annotation of \(\Areg\),
together with \(\mathrm{ctx}_4\), the model predicts the consequence along
the queried Effect channel for the specified recipient. We write this query
as $E_q(r)$, where
$q\in\{\mathrm{men},\mathrm{aff},\mathrm{phy}\}$ identifies the channel
and $r$ the recipient:
\taskformulation{eq:task4_mapping}{
(\mathrm{ctx}_4;\,\Astate,\Amanifest,[\Areg])
\longrightarrow
Q_4=\{E_q(r)\}
}
Unlike Task~3, the supplied \(\Astate\) also includes the fine-grained
description. \(\Amanifest\) and \(\Areg\) follow the same provision rules
as in Task~3. Task~4 infers the downstream consequence for the specified
recipient along the queried Effect channel.

\tasktitle{Task 5. Full Chain Reconstruction}
Task~5 jointly recovers the complete Affective Blueprint from
\(\mathrm{ctx}_5\), which identifies the focal subject, affective object,
and episode anchor. The query specifies the Condition and Affect
constructs together with the requested Effect channels and their recipients:
\taskformulation{eq:task5_mapping}{
(\mathrm{ctx}_5)
\longrightarrow
Q_5=\{\Cext,[\Cint],\Astate,\Amanifest,[\Areg],E\}
}
\end{subequations}

Here $E$ collectively denotes the requested $E_q(r)$ entries. The affective
object is given in \(\mathrm{ctx}_5\) and is neither inferred nor scored.
No other Blueprint fields are supplied.
Question and answer templates appear in Appendix~C.4.

\begin{figure*}[!t]
  \centering
    \includegraphics[width=\linewidth]{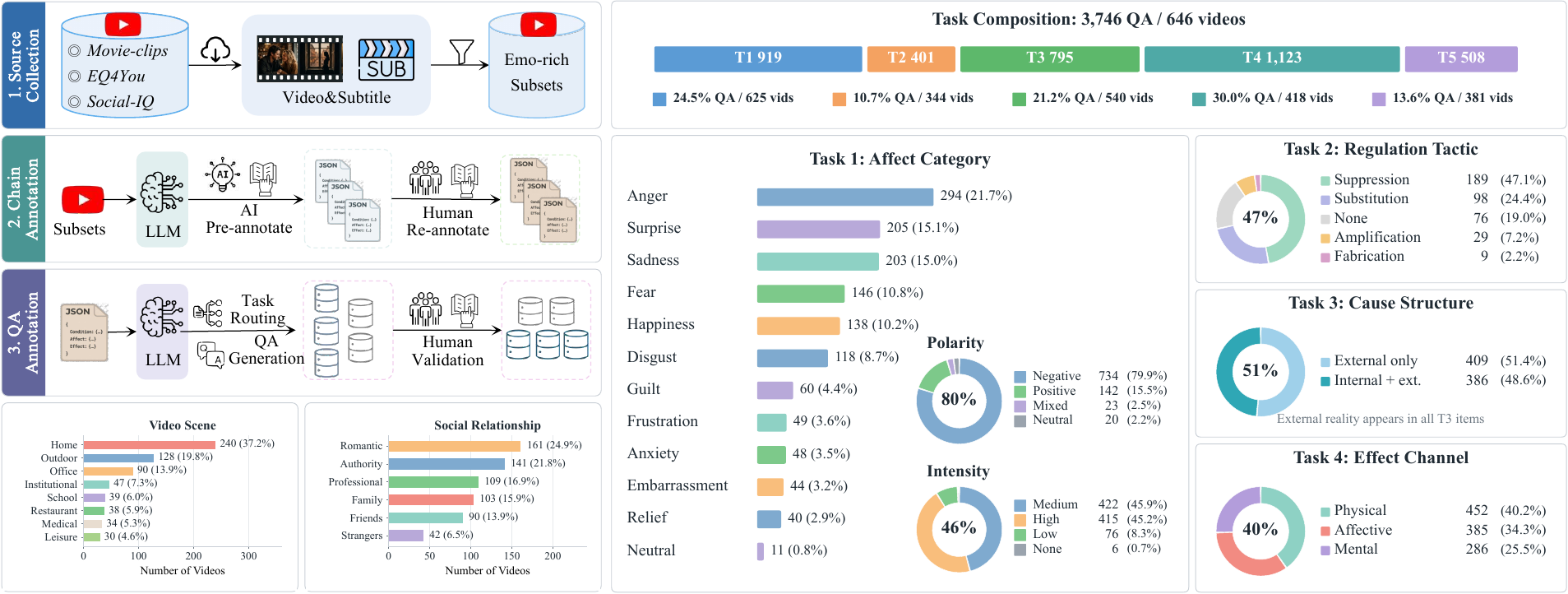}
    \caption{\benchmark{} construction pipeline and dataset statistics,
    with the three-stage construction workflow on the left.}
    \label{fig:annotation_pipeline_and_statistics}
\end{figure*}

\subsection{Data Construction}
\label{sec:data}

\datastep{1}{Source Collection}
We construct \benchmark{} from social video clips drawn from
Movie-clips~\cite{christophschuhmann2024movieclips},
EQ4You~\cite{eq4you2024movieclips}, and
Social-IQ~\cite{zadeh2019socialiq}. The raw collections are first screened
by an MLLM that summarizes each clip's basic content and filters out clips
that lack sufficient context for a recognizable affective episode, such as isolated reaction shots,
exposition-only dialogue, single-person monologues, and highly stylized
performances. Clips that pass this initial screening are further verified
during chain annotation, where additional unsuitable
clips are discarded. This two-stage process yields a curated pool of
context-rich social interaction clips.

\datastep{2}{Chain Annotation}
The basic annotation unit is a subject-centered affective chain, namely
a subject-centered affective episode anchored at a specific moment in the
video. We adopt this granularity because the Affective Blueprint is defined around a single
subject's process from Condition through Affect to Effect. Each chain is
stored as a structured JSON object organized by the three Blueprint stages.
A video may yield multiple chains across characters or moments, each
localized by an episode anchor such as an utterance, facial display, body
action, or vocal cue.

For each selected clip, an MLLM drafts candidate chains from the video and
subtitles. Human annotators then reject invalid proposals, correct
inaccurate fields, add missed episodes, and verify the retained content
against the subject, anchor, and scene evidence. Appendix~C.1
reports field-level rewrite rates for retained candidate chains.
During verification, a field is filled only when the video and subtitles
provide sufficient evidence for it. A chain is retained as long as it
contains sufficient information for at least one valid stage or cross-stage
relation. The independent re-annotation protocol and agreement results are
reported in Appendix~C.2, while guidelines for selected key fields are
provided in Appendix~C.3.

\datastep{3}{QA Annotation}
A verified chain may support multiple evaluation tasks, depending on which
parts of the Blueprint are available in its annotation. We therefore assign
each chain only to the tasks it can support and convert the corresponding
annotations into task-specific QA pairs through predefined templates. For
each eligible task, the template determines which information is exposed as
the episode context $\mathrm{ctx}_t$ and Known Constructs $K_t$, which
constructs are queried as $Q_t$, and how the relevant chain fields are
rewritten into the question and reference answer. The same chain may
consequently yield multiple QA pairs under different task templates. Each
resulting pair undergoes final human validation against the original chain
annotation.

\datatitle{Dataset Statistics}
\benchmark{} contains \TraceTotalQAs{} structured QA pairs from
\TraceTotalVideos{} videos. The videos span settings such as homes and
workplaces, covering close personal relationships as well as professional
and authority-based interactions.

Task~1 uses a 12-category affective vocabulary comprising basic categories
grounded in Ekman's account~\cite{ekman1992basic}, additional categories such
as guilt and relief, and Neutral. Task~2 covers all five regulation tactics,
from intensity modulation and category substitution to None, where the
display matches the felt state. Task~4 contains between \TraceEffectMentalCount{}
and \TraceEffectPhysicalCount{} queries for each of the three Effect
channels. Fig.~\ref{fig:annotation_pipeline_and_statistics} summarizes
task composition and annotation distributions.

\subsection{Evaluation Protocol}
\label{sec:bench:evaluation}

The five tasks require predictions of both categorical labels and
open-ended descriptions, which we score at the field level before
aggregating them into task scores. Categorical predictions are compared
directly with the reference: since an Affective State can contain multiple
categories, category sets are scored by set-F1, and single-label fields by
exact match.

Open-ended descriptions can express equivalent content in different
wording, so we use DeepSeek-V4-Flash to rate each prediction against the
annotated reference. Semantic adequacy, $d_1$, measures whether the
prediction captures the intended content for the correct character and
episode. A prediction may also overlap with the reference by repeating
generic statements or listing competing guesses, and a separate redundancy
rating, $d_2$, estimates how much of the answer adds no information relevant
to the queried field. Each field is rated repeatedly, and the $d_1$ and
$d_2$ values are averaged separately, giving the field score
\begin{equation}
\label{eq:field_score}
s(f)=\bar d_1(f)\bigl(1-\lambda\,\bar d_2(f)\bigr),
\end{equation}
where both ratings lie in $[0,1]$ and $\lambda$ bounds the redundancy
discount. We set $\lambda=0.5$, so that redundancy can reduce a field score
by at most half while semantic adequacy remains its basis. Appendix~B.1
provides the field-matching rules and judging rubric.

Field scores are combined along the structure of the Affective Blueprint:
fields are first averaged within constructs and then constructs within
stages, so that a construct described by several fields, such as
Manifestation, does not outweigh the others. Full-chain reconstruction thus
gives Condition, Affect, and Effect equal weight. In Task~2, Regulation
Evidence is scored alongside the applicable Regulation fields, with equal
weight for each. With $N_t$ evaluated items
in task $t$, $F_{ti}$ the evaluated fields of item $i$, and $w_{ti}(f)$
their normalized task-specific weights, the overall score averages
item scores within each task and then across tasks:
\begin{equation}
\label{eq:task_score}
\TRACEScore
=\frac{100}{5}\sum_{t=1}^{5}\frac{1}{N_t}
\sum_{i=1}^{N_t}\sum_{f\in F_{ti}}w_{ti}(f)\,s_{ti}(f).
\end{equation}
Appendix~B.2 specifies task-dependent scoring conditions and the
treatment of missing fields, Appendix~B.3 the scoring and reporting of unparsed
outputs, and Appendix~A.7 human solvability and judge calibration.

We use video-level cluster bootstrap (10{,}000 resamples), sampling videos
with replacement and retaining all their items, with shared resamples
across models and tasks. We apply Holm correction to comparisons against
14 baselines on each of the five tasks. The same resamples yield the
95\% confidence interval for the five-task average margin over GPT-5 (thinking).

\section{\method{}: Cognition-Grounded Structured Reasoning}
\label{sec:method}
\label{sec:method:motivation}

\begin{figure*}[t]
  \centering
  \includegraphics[width=0.96\textwidth]{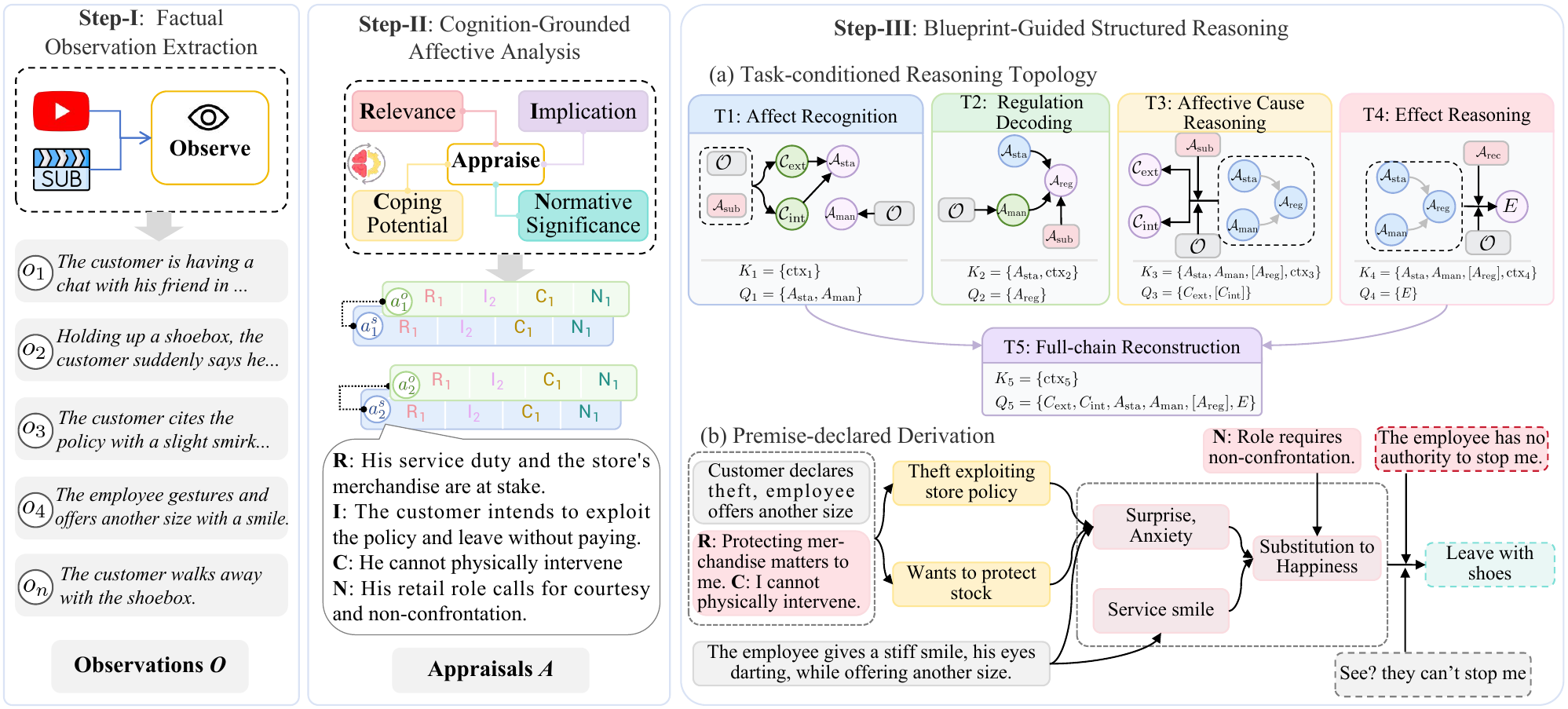}
  \caption{Overview of \method{}. Factual observations are interpreted through
  character-centered appraisal and used in Blueprint-guided derivation.
  (a) Gray arrows denote intrinsic relations, while black arrows specify
  inference dependencies and premise requirements. Blue, green, and purple
  construct nodes denote known, intermediate,
  and queried constructs, respectively.
  (b) An abridged shoe-store example illustrates how observations, appraisals,
  and intermediate conclusions jointly support subsequent inferences.}
  \Description{Three-step TRACER pipeline with task-conditioned reasoning
  topologies and an abridged shoe-store derivation example.}
  \label{fig:method_overview}
\end{figure*}

TRACE tasks involve heterogeneous Blueprint constructs, including observable
manifestations, psychological drivers, affective states, regulation, and
downstream effects. Inferring these constructs requires different forms of
evidence and attention to their dependencies within an affective episode.
This complexity presents two challenges to current models. \textbf{First}, models must
interpret the affective meaning behind the observed behaviors and
interactions. They must determine whether an event functions as an affective
trigger for the subject and whether an outward manifestation expresses the
felt state or a regulated display. Reliance on visible cues leads to a surface
reading that collapses these alternatives. \textbf{Second}, placing heterogeneous
constructs in a single generation process increases factual drift, especially in
multi-field, long-chain reconstruction. An upstream interpretation may be
promoted into an established fact and then used to generate a plausible
downstream event beyond the recorded evidence. The resulting chain remains
linguistically coherent while departing from what occurs in the video.

\looseness=-1
To address these challenges, we propose \method{}, which constrains every
inference to premises from factual observations, their affective meanings,
and relevant Blueprint constructs.

\subsection{Method Overview}
\label{sec:method:overview}

As shown in Fig.~\ref{fig:method_overview}, \method{} transforms multimodal
social video into task-specific predictions through three stages.
\textit{Factual Observation Extraction} converts the video and aligned
subtitles into an indexed record $\mathcal O$ of observable events,
behaviors, and utterances. \textit{Cognition-Grounded Affective Analysis}
interprets the relevant observations from the involved characters'
perspectives, producing character-specific appraisal readings $\mathcal A$.
\textit{Blueprint-Guided Structured Derivation} combines these observations
and appraisal readings with the task's known and queried constructs to
derive the requested outputs through premise-grounded reasoning.

The five tasks share this pipeline and the \blueprint{}, while the final
derivation is organized according to each task's supplied and queried
constructs. All three stages use stage-specific prompts with off-the-shelf
MLLMs or LLMs and require no additional training.

\subsection{Factual Observation Extraction}
\label{sec:method:observation}

The first stage establishes the factual record used throughout reasoning.
Given a complete video $v$ and aligned subtitles $s$, the observation
module (Appendix~A.2) produces
\begin{equation}
  \mathcal O=\Observe(v,s)
  =\{o_i\}_{i=1}^{n},
  \label{eq:observation}
\end{equation}
\looseness=-1
where each observation item $o_i$ records factual content from a local video
interval or its aligned subtitles and receives a stable identifier.
Together, they form a temporally ordered account of the participants, actions,
interactions, facial and bodily behavior, vocal delivery, and spoken content in
the scene.

This stage records perceptible content before assigning its affective meaning.
Affective state, internal driver, regulation, and downstream effect are
therefore reserved for later reasoning. This boundary gives every subsequent
conclusion an independent factual source and keeps an inferred state from
becoming evidence for its own manifestation.

The model uses the task context $\mathrm{ctx}_t$ as an anchor to locate
relevant observations in $\mathcal O$. It then selects and groups observation
items $S_t\subseteq\mathcal O$ for appraisal and reasoning. The full record
remains available for citation throughout the subsequent stages.

\subsection{Cognition-Grounded Affective Analysis}
\label{sec:method:appraisal}

An observed event acquires affective meaning through its relation to a
character's goals, expectations, resources, and social position. Inspired by
cognitive appraisal theory~\cite{scherer2001appraisal}, we analyze the
selected items from the position of the character involved. The analyzed
characters are specified by the task, namely the subject and, when the
query includes an Effect, the target recipient. For character
$r$, a selected group of items $S_{t,k}\subseteq S_t$, and appraisal
dimension $d$, the analysis produces
\begin{equation}
  a_{r,k}^{(d)}
  =\Appraise_d\!\left(S_{t,k}\mid r\right),
  \qquad d\in\mathbb D,
  \label{eq:appraisal}
\end{equation}
where $a_{r,k}^{(d)}$ is a complete natural-language proposition stating what
the cited facts mean from $r$'s position. Its evidence links remain attached
to the proposition. The four dimensions in $\mathbb D$ examine complementary
parts of this meaning:

\begin{tcolorbox}[
  enhanced,
  colback=theorybluebg,
  frame hidden,
  borderline west={1pt}{0pt}{theoryblue},
  title={Cognitive Appraisal Dimensions},
  breakable,
  before upper={\interlinepenalty=10000},
  coltitle=theoryblue,
  fonttitle=\bfseries\small,
  boxrule=0pt,
  arc=0pt,
  left=3pt,
  right=4pt,
  top=2pt,
  bottom=2pt,
  before skip=3pt,
  after skip=4pt
]
\begin{itemize}[leftmargin=*,itemsep=1pt,parsep=0pt,topsep=0pt]
  \item \textbf{Relevance} identifies why the event attracts the character's
  attention through novelty, intrinsic pleasantness, or relation to an active
  goal or need.
  \item \textbf{Implication} determines the event's consequences for that goal
  or need through agency, expected outcome, goal conduciveness, and urgency.
  \item \textbf{Coping Potential} examines the character's control, available
  power and resources, and capacity to adapt to the consequences.
  \item \textbf{Normative Significance} evaluates compatibility with the
  character's self-concept, personal standards, and recognized social or moral
  standards.
\end{itemize}
\end{tcolorbox}

The output of each dimension is an evidence-grounded interpretation rather
than a score, level, category, or polarity. We collect the resulting readings
as $\mathcal A$. In Fig.~\ref{fig:method_overview}(a),
$\mathcal A_{\mathrm{sub}}$ and $\mathcal A_{\mathrm{rec}}$ denote the
subject-side and recipient-side readings in this collection.
Subject-side readings are produced for every selected group. When the
queried construct includes an Effect, recipient-side readings are
additionally produced for the group that contains the subject's described
behavior. The same observation can thus
receive different meanings under the subject and recipient perspectives while
each interpretation remains bound to its role and cited facts. These readings
enter Blueprint-guided derivation as cognitive premises, not final field
values. Core prompt instructions appear in Appendix~A.3.

\subsection{Blueprint-Guided Structured Derivation}
\label{sec:method:derivation}

Given the factual observations $\mathcal O$ and appraisal readings
$\mathcal A$ established in the previous stages, the final stage derives the
Blueprint constructs required by the task. The five tasks share the same
affective structure but differ in their known constructs $K_t$ and queried
constructs $Q_t$, requiring different derivation paths. A task-conditioned
reasoning topology specifies which constructs must be established, what
information each inference requires, and their derivation order.
Premise-declared derivation then uses concrete premises from the current
video to carry out these inferences.

\runinhead{Task-conditioned reasoning topology (RT).}\enskip
The \blueprint{} provides the common construct system and the intrinsic
relations among its components. RT organizes the dependencies needed to
establish the queried constructs, introducing missing constructs as
intermediate conclusions when needed. Fig.~\ref{fig:method_overview}(a)
distinguishes intrinsic Blueprint relations in gray from active inference
links in black. The former describe how constructs are related within an
affective episode, while the latter specify how available information is
used to derive a construct for the current task. In Task~3, for example,
supplied Affect constrains the inference of its antecedent Condition,
while the underlying affective process still follows Condition
$\rightarrow$ Affect.

\runinhead{Nodes.}\enskip
We retain External, Internal, State, Manifestation, Regulation, and Effect
as construct nodes. Their roles depend on the task: a construct may be
supplied in $K_t$, introduced as an intermediate conclusion, or queried in
$Q_t$. Premise-source nodes represent observations $\mathcal O$ and
subject-side and recipient-side appraisal collections
$\mathcal A_{\mathrm{sub}}$ and $\mathcal A_{\mathrm{rec}}$.

\runinhead{Links.}\enskip
A construct-dependency link $u\rightarrow x$ requires information from $u$
to infer $x$. External and Internal jointly support State, reflecting how
an external event acquires affective significance in relation to the
subject's internal stance. State and Manifestation jointly support
Regulation through their comparison. Effect draws on the supplied or
derived Affect constructs, including Regulation when applicable.
The topology also specifies the premise sources used by each inference.
External and Internal draw on observations together with subject-side
appraisal. Manifestation is derived directly from observations,
independently of the inferred State. Regulation uses subject-side
appraisal, while Effect uses appraisal from the queried recipient.

We represent the resulting topology as
\begin{equation}
  \mathcal T_t=(V_t,E_t,\prec_t).
  \label{eq:projection}
\end{equation}
where $V_t$ contains the relevant construct and premise-source nodes,
$E_t$ contains the active inference links, and $\prec_t$ places construct
premises before conclusions that use them. Supplied values in $K_t$
remain fixed during derivation.

\runinhead{Premise-declared derivation (PDD).}\enskip
Following the topology, PDD selects concrete observations and appraisal
readings for the current video, together with
supplied constructs or earlier conclusions, and cites them as premises.
This design follows the broader principle of making intermediate reasoning
structure explicit during inference, as explored in prior work on faithful
and structured logical reasoning~\cite{xu2024symbcot,xu2025aristotle}.
For each intermediate or queried construct, the model jointly produces
the construct identity, its premise references, and the inferred field values:
\begin{equation}
  c_j=\langle x_j,P_j,y_j\rangle.
  \label{eq:node}
\end{equation}
Here $x_j$ identifies the construct, $P_j$ records its premise references,
and $y_j$ contains its inferred field values. The derivation sequence
$\mathcal D_t=(c_1,\ldots,c_L)$ follows $\prec_t$. Established nodes become
available to later steps with fixed identifiers and field values, while
the full observation and appraisal collections remain accessible for
premise selection. The final answer
values $\mathcal D_t[Q_t]$ are extracted from the queried nodes rather
than generated separately. These form the task answer $\widehat Y_t$,
with Task~2 additionally including Regulation Evidence $M_{\mathrm{reg}}$.

In Fig.~\ref{fig:method_overview}(b), the customer's declaration and the
employee's appraisal establish External, while the employee's concerns
about merchandise and non-confrontation support Internal. These
intermediate conclusions then support the inferred State. For Task~1,
the Condition constructs therefore serve as intermediate premises even
though the task queries State and Manifestation.

\section{Experiments}
\label{sec:experiments}

\newcommand{\mainresultstable}{%
\begin{figure*}[!t]
\centering
\begin{minipage}[t]{0.635\textwidth}
\vspace{0pt}          %
\centering
\captionof{table}{Main results on \benchmark{}. Avg. averages T1--T4 and
T5 Full. Model scores use the main evaluation set; Human scores use a subset
of 50 questions per task. Matched model scores are in Appendix~A.7.
Per-stage T5 scores are in Section~\ref{sec:exp:within_task}.
Unless otherwise stated, tables use \textbf{bold} for the best model results and
\underline{underlining} for the second best. * marks cells with substantial
parsing failures, and $\dag$ cells where most outputs could not be parsed.}
\label{tab:main_results}
\renewcommand{\expnum}[1]{##1}
\footnotesize
\setlength{\tabcolsep}{3.2pt}
\renewcommand{\arraystretch}{1.38}
\begin{tabular*}{\textwidth}{@{\extracolsep{\fill}}l cccc c c@{}}
\toprule
\textbf{Model} & \textbf{T1 Rec.} & \textbf{T2 Reg.} &
\textbf{T3 Cause} & \textbf{T4 Effect} & \textbf{T5 Full} &
\textbf{Avg.} \\
\midrule
Human ($n=50$) & 83.60 & 72.32 & 85.55 & 78.90 & 75.40 & 79.15 \\
\midrule
\multicolumn{7}{@{}l}{\textbf{\textit{Affect Specialized}}} \\
Emotion-LLaMA~\cite{cheng2024emotionllama} & \expnum{0.45} & \expnum{5.22}\textsuperscript{*} & \expnum{2.26}\textsuperscript{\dag} & \expnum{0.31}\textsuperscript{\dag} & \expnum{0.20}\textsuperscript{\dag} & \expnum{1.69} \\
AffectGPT~\cite{lian2025affectgpt} & \expnum{34.57} & \expnum{4.64}\textsuperscript{\dag} & \expnum{17.99}\textsuperscript{*} & \expnum{15.79}\textsuperscript{*} & \expnum{20.87}\textsuperscript{*} & \expnum{18.77} \\
Emotion-Qwen & \expnum{29.45}\textsuperscript{*} & \expnum{24.24}\textsuperscript{*} & \expnum{21.04}\textsuperscript{*} & \expnum{13.49} & \expnum{19.13}\textsuperscript{*} & \expnum{21.47} \\
\midrule
\multicolumn{7}{@{}l}{\textbf{\textit{Open Source Generic}}} \\
LLaVA-OneVision-7B~\cite{li2024llavaov} & \expnum{35.13} & \expnum{23.51} & \expnum{25.28} & \expnum{4.11} & \expnum{5.47}\textsuperscript{\dag} & \expnum{18.70} \\
LLaVA-NeXT-Video-32B~\cite{li2024llavanext} & \expnum{25.89} & \expnum{27.01} & \expnum{27.13}\textsuperscript{*} & \expnum{23.61} & \expnum{13.80}\textsuperscript{*} & \expnum{23.49} \\
MiniCPM-V-4.5~\cite{yao2025minicpmv45} & \expnum{37.30} & \expnum{29.19} & \expnum{43.04} & \expnum{28.27} & \expnum{29.43} & \expnum{33.45} \\
GLM-4.6V-Flash-9B~\cite{glm2025glm41v} & \expnum{40.80} & \expnum{28.85} & \expnum{42.30} & \expnum{28.81} & \expnum{32.36} & \expnum{34.62} \\
LLaVA-OneVision-70B~\cite{li2024llavaov} & \expnum{42.74} & \expnum{26.73} & \expnum{47.30} & \expnum{33.03} & \expnum{34.62} & \expnum{36.88} \\
Qwen3-Omni-30B~\cite{qwen2025qwen3omni} & \expnum{46.28} & \expnum{33.17} & \expnum{51.69} & \expnum{39.81} & \expnum{41.78} & \expnum{42.55} \\
InternVL3.5-38B~\cite{chen2025internvl35} & \expnum{45.35} & \expnum{28.05} & \expnum{46.16} & \expnum{32.19} & \expnum{37.07} & \expnum{37.76} \\
Qwen3-VL-32B~\cite{qwen2025qwen3vl} & \expnum{50.97} & \expnum{35.49} & \expnum{60.42} & \expnum{44.38} & \expnum{50.16} & \expnum{48.28} \\
\midrule
\multicolumn{7}{@{}l}{\textbf{\textit{Closed Source Generic}}} \\
GPT-5 (non-thinking) & \expnum{56.13} & \expnum{37.26} & \expnum{70.74} & \expnum{52.36} & \expnum{58.22} & \expnum{54.94} \\
Gemini-3-Pro & \expnum{54.32} & \expnum{40.55} & \expnum{70.74} & \expnum{51.48} & \expnum{56.05} & \expnum{54.63} \\
GPT-5 (thinking) & \expnum{\underline{56.61}} & \expnum{\underline{41.56}} & \expnum{\underline{77.11}} & \expnum{\underline{53.75}} & \expnum{\underline{60.79}} & \expnum{\underline{57.96}} \\
\midrule
\method{} (Ours) & \expnum{\textbf{61.67}} & \expnum{\textbf{49.53}} & \expnum{\textbf{84.55}} & \expnum{\textbf{59.47}} & \expnum{\textbf{63.59}} & \expnum{\textbf{63.76}} \\
\bottomrule
\end{tabular*}
\end{minipage}\hfill
\begin{minipage}[t]{0.355\textwidth}
\vspace{0pt}
\begin{minipage}[t]{0.96\linewidth}
\captionsetup[subfloat]{farskip=3pt,captionskip=3pt}
\vspace{-0.45em}
\centering
\includegraphics[width=0.92\linewidth]{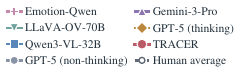}\par
\vspace{-0.15em}
\centering
\subfloat[Component profile.]{%
\includegraphics[width=0.75\linewidth]{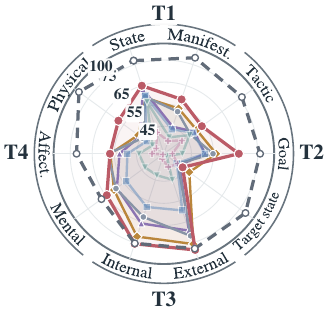}
\label{fig:per_task_radar}}\par
\vspace{0.05em}
\centering
\subfloat[Family preference.]{%
\includegraphics[width=0.775\linewidth]{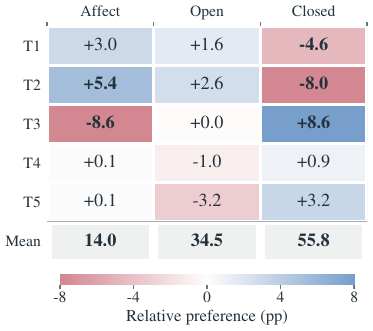}
\label{fig:model_family_preference}}\par
\end{minipage}
\caption{Model profiles by (a) task component and (b) family.
Human scores average two participants.}
\label{fig:main_side}
\end{minipage}
\end{figure*}
}

\mainresultstable

\subsection{Experimental Setup}
\label{sec:exp:setup}

\exphead{Model groups}
We compare \method{} with the three baseline groups listed in
Table~\ref{tab:main_results}. \textit{Open-source generic models} cover different
scales and multimodal architectures, including
LLaVA-OneVision~\cite{li2024llavaov}, MiniCPM-V-4.5~\cite{yao2025minicpmv45},
InternVL3.5~\cite{chen2025internvl35}, and Qwen3-VL~\cite{qwen2025qwen3vl}.
\textit{Affective-specialized models} are trained specifically for emotion
understanding and include Emotion-LLaMA~\cite{cheng2024emotionllama},
AffectGPT~\cite{lian2025affectgpt}, and Emotion-Qwen~\cite{huang2025emotion}.
\textit{Closed-source models} comprise GPT-5~\cite{singh2025openai}, evaluated
in thinking and non-thinking modes, and Gemini-3-Pro~\cite{deepmind2026gemini3pro}.
{These groups allow comparison of model-family strengths and limitations
across TRACE tasks.}

\exphead{Implementation details}
{Baseline models receive video frames, available subtitles appended to the
prompt, and the task question. Only Qwen3-Omni-30B receives audio.
Frame budgets are model-specific: GPT-5 uses eight uniformly sampled frames,
while Qwen3-VL-32B uses its video processor with a 60-frame budget, both
covering the full clip. The question provides known information and a JSON
template specifying the requested fields. For \method{}, Qwen3-VL generates
factual observation records, and GPT-5 generates character-specific appraisals
and subsequent construct-level inferences. Appendix~A.1 gives full inference
settings. All models follow the evaluation protocol in
Section~\ref{sec:bench:evaluation}.}

\subsection{Main Results on \benchmark{}}
\label{sec:exp:main}

\exphead{Overall performance}
{Closed-source general-purpose models lead the baselines in
Table~\ref{tab:main_results}, with GPT-5 (thinking) scoring highest overall.}
\method{} outperforms every model baseline on each task
($p<0.05$, bootstrap test, Holm-corrected).
Its average margin over GPT-5 (thinking) is \expnum{5.80} points (bootstrap 95\% CI
\expnum{4.42} to \expnum{7.17}). The largest gains occur in
T2 regulation decoding (\expnum{7.97} points) and T3 cause diagnosis
(\expnum{7.44} points), followed by T4 effect reasoning and T1 affect
recognition. {Gains span recognition and reasoning.
Full-chain reconstruction shows the smallest improvement.}
Despite this improvement, T2 remains the lowest-scoring task for both
GPT-5 (thinking) and \method{}.

{For a matched human--model comparison, we evaluate models on the same
50 questions per task answered by the two human participants (Appendix~A.7).
On this subset, humans average \expnum{79.15}, GPT-5 (thinking) \expnum{58.75}, and
\method{} \expnum{63.58}, leaving a human--\method{} gap of
\expnum{15.57} points. \method{} is closest to humans on T3:
\expnum{84.39} versus \expnum{85.55}, a \expnum{1.16}-point gap.
The other four tasks show larger gaps.}

\exphead{Scale and thinking yield task-selective gains}
Within LLaVA-OneVision, scaling from 7B to 70B improves all five tasks,
with the largest gains in cause reasoning, effect reasoning, and full-chain
reconstruction. However,
Qwen3-VL-32B outperforms the 70B model across all tasks, showing that
parameter count alone does not explain performance differences across models.
Enabling thinking in GPT-5 primarily benefits cause diagnosis and
regulation decoding, with smaller gains in affect recognition and effect
reasoning. For example, T3 improves by \expnum{6.37} points, whereas T1
changes by only \expnum{0.48} points.

\exphead{Model families favor different tasks}
Fig.~\ref{fig:model_family_preference} shows distinct task profiles across
the three model families. Affect-specialized models exhibit
relative strengths in T1 affect recognition and T2 regulation decoding,
but a pronounced weakness in T3 cause reasoning. Open-source general-purpose
models have a more balanced profile across individual tasks, yet show a
deficit in T5, where Condition, Affect, and Effect must be reconstructed
jointly. Closed-source models display a contrasting pattern: their strengths
are concentrated in cause reasoning and full-chain reconstruction, while
T1 and T2 are comparatively weaker parts of their profile.
{Scores are centered by family and task averages (supplementary Eq.~S1),
so these profiles describe within-family task preferences, not absolute
rankings across families. Closed-source models can therefore be relatively
weaker on T1 and T2 yet score higher than affect-specialized models.}

\subsection{Evidence Ablation}
\label{sec:exp:evidence_ablation}

\exphead{Design}To assess the contribution of video evidence beyond the information supplied
in each task, we evaluate GPT-5 (thinking) and Qwen3-VL-32B with the question
and supplied fields alone, then add subtitles, frames, or both.
Frame inputs use eight frames for GPT-5 and 60 for Qwen.
We also test GPT-5 with subtitles and 16 frames.

\begin{figure}[!t]
  \centering
  \setlength{\abovecaptionskip}{5pt}
  \includegraphics[width=\columnwidth]{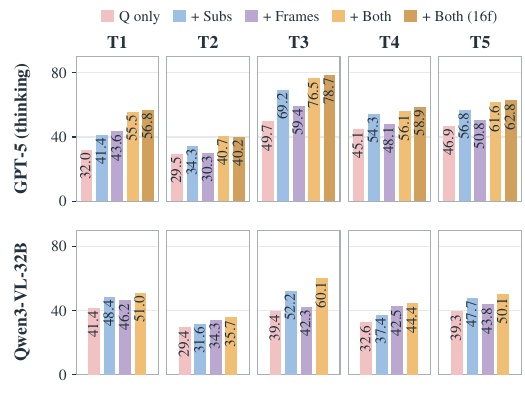}
  \caption{Evidence ablation. Q includes the question and supplied fields.
  Frames denotes 8 frames for GPT-5 and 60 for Qwen.
  Both (16f): GPT-5 with subtitles and 16 frames.}
  \label{fig:evidence_ablation}
\end{figure}

\begin{figure}[!t]
  \vspace{0pt}\centering
  \setlength{\abovecaptionskip}{5pt}
  \includegraphics[width=\linewidth,trim=0 25bp 0 0,clip]{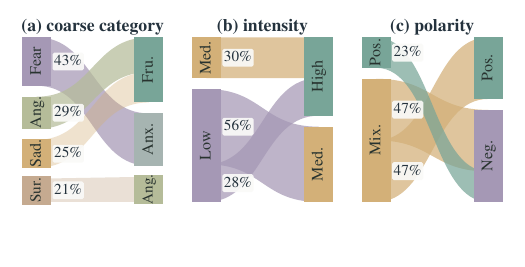}
  \caption{T1 confusions from true (left) to predicted (right) labels.
  Band widths show proportions within each true class.}
  \label{fig:t1_state}
\end{figure}

\exphead{Results and analysis}Joint subtitle and frame input outperforms
question-only input on all five tasks for both models
(Fig.~\ref{fig:evidence_ablation}), showing that the supplied fields do not
replace episode-specific evidence. Neither channel alone matches joint
input on any task, supporting the need for multimodal evidence in affective
understanding. Channel preferences nevertheless vary: both models favor
subtitles on T3 and T5, but their preferences differ on T1, T2, and T4.
For example, Qwen favors frames on T4, whereas GPT-5 favors subtitles.
Increasing GPT-5's frame count from eight to 16 produces small gains on
four tasks and a slight decrease on T2. Combining evidence channels
therefore provides a more consistent benefit than denser visual sampling.

\setcounter{topnumber}{2}
\begin{table}[t]
  \centering
  \caption{T1 field-level performance. Shaded rows report construct scores
  aggregated within each instance.}
  \label{tab:t1_fields}
  \scriptsize
  \setlength{\tabcolsep}{1.5pt}
  \renewcommand{\arraystretch}{1.10}
  \begin{tabularx}{\linewidth}{@{}l>{\hsize=0.92\hsize\raggedleft\arraybackslash}X>{\hsize=0.98\hsize\raggedleft\arraybackslash}X>{\hsize=0.92\hsize\raggedleft\arraybackslash}X>{\hsize=1.06\hsize\raggedleft\arraybackslash}X>{\hsize=0.92\hsize\raggedleft\arraybackslash}X>{\hsize=1.00\hsize\raggedleft\arraybackslash}X>{\hsize=1.20\hsize\raggedleft\arraybackslash}X@{}}
    \toprule
    & \textbf{EmoQ*} & \textbf{OV-70B} & \textbf{Q3-VL} & \textbf{GPT5-N} & \textbf{Gem-3} & \textbf{GPT5-T} & \textbf{TRACER} \\
    \midrule
    \rowcolor{black!12}
    \textbf{\textit{Affective state}} & 36.8 & 57.3 & 60.1 & 59.7 & \underline{63.3} & 59.2 & \textbf{64.9} \\
    \quad Category & 27.2 & 40.1 & 38.8 & 36.1 & \textbf{48.1} & \underline{41.5} & 40.5 \\
    \quad Description & 11.7 & 36.8 & 48.1 & 55.6 & 53.7 & \underline{57.3} & \textbf{60.0} \\
    \quad Polarity & 57.5 & \underline{80.1} & \textbf{81.4} & 69.1 & \textbf{81.4} & 71.3 & 77.5 \\
    \quad Intensity & 35.4 & 61.1 & 64.2 & \underline{64.5} & 58.2 & 63.3 & \textbf{66.5} \\
    \quad Object & 51.8 & 67.7 & 67.8 & 73.0 & \underline{74.7} & 62.5 & \textbf{79.9} \\
    \addlinespace[1.5pt]
    \rowcolor{black!12}
    \textbf{\textit{Manifestation}} & 22.1 & 28.2 & 41.8 & 52.6 & 45.4 & \underline{54.0} & \textbf{58.4} \\
    \quad Facial & 30.0 & 32.9 & 47.2 & 56.9 & 48.5 & \textbf{59.2} & \underline{57.7} \\
    \quad Body & 20.4 & 23.1 & 36.6 & 46.2 & 37.3 & \underline{47.2} & \textbf{55.6} \\
    \quad Verbal & 16.0 & 31.0 & 44.8 & 60.1 & 55.4 & \underline{61.6} & \textbf{70.6} \\
    \quad Vocal & 23.3 & 27.5 & 43.0 & 52.4 & 45.6 & \underline{54.2} & \textbf{56.6} \\
    \bottomrule
  \end{tabularx}
\end{table}

\begin{table*}[!t]
\renewenvironment{table}[1][]{}{}
\begin{minipage}[t]{\columnwidth}
\vspace{0pt}
\begin{table}[t]
  \centering
  \caption{T2 field-level performance. Goal and Regulation Evidence use regulated instances. Target state is scored only when its reference category differs from the source category.}
  \label{tab:t2_fields}
  \scriptsize
  \setlength{\tabcolsep}{2.0pt}
  \renewcommand{\arraystretch}{1.12}
  \begin{tabular*}{\linewidth}{@{\extracolsep{\fill}}lrrrrrrr@{}}
    \toprule
    & \textbf{EmoQ*} & \textbf{OV-70B} & \textbf{Q3-VL} & \textbf{GPT5-N} & \textbf{Gem-3} & \textbf{GPT5-T} & \textbf{TRACER} \\
    \midrule
    Tactic & 36.7 & 46.9 & 48.9 & 47.4 & 49.5 & \underline{50.9} & \textbf{53.8} \\
    Target state & 28.0 & 5.6 & 26.2 & 28.0 & 31.6 & \textbf{46.7} & \underline{37.7} \\
    Goal & 13.5 & 30.1 & 51.1 & 54.8 & 51.2 & \underline{56.2} & \textbf{66.5} \\
    Reg. evidence & 10.0 & 8.7 & 18.1 & \underline{23.6} & 18.1 & \underline{23.6} & \textbf{50.3} \\
    \bottomrule
  \end{tabular*}
\end{table}

\end{minipage}\hfill
\begin{minipage}[t]{\columnwidth}
\vspace{0pt}
\begin{table}[t]
  \centering
  \caption{T4 effect-channel performance, including self-directed and other-directed effects. Each channel is scored on its applicable instances.}
  \label{tab:t4_fields}
  \scriptsize
  \setlength{\tabcolsep}{2.0pt}
  \renewcommand{\arraystretch}{1.12}%
  \begin{tabular*}{\linewidth}{@{\extracolsep{\fill}}lrrrrrrr@{}}
    \toprule
    & \textbf{EmoQ*} & \textbf{OV-70B} & \textbf{Q3-VL} & \textbf{GPT5-N} & \textbf{Gem-3} & \textbf{GPT5-T} & \textbf{TRACER} \\
    \midrule
    Mental & 2.1 & 38.1 & 53.5 & 59.4 & 58.8 & \underline{60.5} & \textbf{64.5} \\
    \addlinespace[0.56\normalbaselineskip]
    Affective & 19.6 & 40.7 & 49.0 & \textbf{57.8} & 52.6 & 56.8 & \underline{57.0} \\
    \addlinespace[0.56\normalbaselineskip]
    Physical & 15.5 & 23.3 & 34.6 & 43.3 & 45.9 & \underline{46.9} & \textbf{58.0} \\
    \bottomrule
  \end{tabular*}
\end{table}

\end{minipage}
\end{table*}

\subsection{Within-Task Performance Breakdown}
\label{sec:exp:within_task}

Task averages can mask uneven component performance. Components are scored
over applicable instances, with full results in Appendix~A.4.

\movetitle{Task~1. Grounded Affect Recognition}
Table~\ref{tab:t1_fields} shows a consistent gap between the two first-order
components: State scores higher than Manifestation for all six baselines,
with larger gaps for the two open-source models than for GPT-5. Enabling
thinking improves Manifestation by \expnum{1.4} points, while State remains
nearly unchanged. Looking more closely at the State predictions reveals
where the remaining errors concentrate. Fig.~\ref{fig:t1_state} shows
recurrent confusions between semantically related categories, particularly
fear and anxiety, and anger and frustration. Intensity predictions
frequently shift both low- and high-intensity states toward medium, while
mixed polarity is often reduced to a single polarity. Thus, higher State
scores still mask errors in category, intensity, and mixed polarity.

\begin{figure}[!t]
  \vspace{0pt}\centering
  \setlength{\abovecaptionskip}{5pt}
  \includegraphics[width=\linewidth]{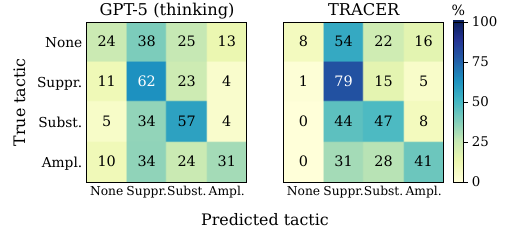}
  \caption{T2 tactic confusion for GPT-5 (thinking) and \method{}.
  Rows: reference; columns: prediction; values: row percentages.
  Fabrication is omitted from both axes.}
  \label{fig:t2_tactic}
\end{figure}

\begin{figure*}[tp]
\begin{minipage}[t]{\columnwidth}
  \vspace{0pt}\centering
  \setlength{\abovecaptionskip}{5pt}
  \parbox[c][165bp][c]{\linewidth}{\centering
  \includegraphics[width=\linewidth,height=165bp,keepaspectratio]{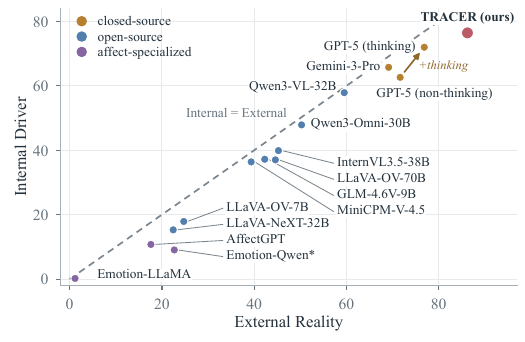}}
  \caption{T3 External Reality versus Internal Driver scores by model.
  The dashed line marks equal scores. The arrow shows GPT-5's change
  with thinking.}
  \label{fig:t3_cause}
\end{minipage}\hfill
\begin{minipage}[t]{\columnwidth}
  \vspace{0pt}\centering
  \setlength{\abovecaptionskip}{5pt}
  \parbox[c][165bp][c]{\linewidth}{\centering
  \includegraphics[width=\linewidth,height=165bp,keepaspectratio]{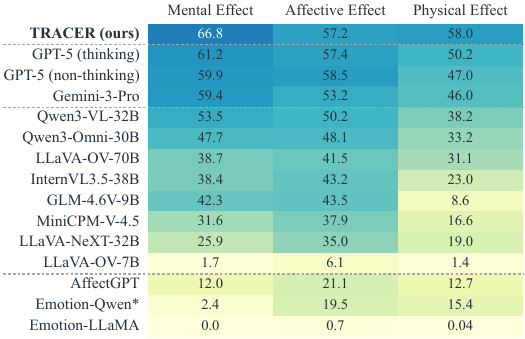}}
  \caption{T4 effect-channel scores on other-directed instances.
  Dashed lines separate \method{}, closed-source, open-source, and
  affect-specialized models.}
  \label{fig:t4_channels}
\end{minipage}
\end{figure*}

\movetitle{Task~2. Regulation Decoding}
For Task~2, the clearest error pattern appears in tactic prediction.
Models frequently infer active regulation when the reference tactic is
None, with false-alarm rates ranging from \expnum{76}\% to \expnum{96}\%
across GPT-5 (thinking), Qwen3-VL, and \method{}.
As shown in Fig.~\ref{fig:t2_tactic}, Suppression is the most common
prediction for these unregulated cases. \method{} identifies Suppression
and Amplification more often than GPT-5 (thinking), but also recognizes
None less often, indicating a stronger tendency to over-detect regulation.
This pattern suggests that tactic prediction remains particularly difficult
at the boundary between genuinely regulated displays and cases without
active regulation. Beyond tactic prediction, thinking substantially
improves Target State for GPT-5, from \expnum{28.0} to \expnum{46.7}
(Table~\ref{tab:t2_fields}), suggesting that reasoning helps infer the
intended target affect given the underlying felt state.

\movetitle{Task~3. Affective Cause Reasoning}
{All evaluated models across the three families score lower on
Internal Driver than External Reality, falling below $y=x$ in
Fig.~\ref{fig:t3_cause}. Thinking improves GPT-5's Internal Driver score
more than External Reality, narrowing this gap. \method{} scores highest
on both components, improving more over GPT-5 (thinking) on External
Reality. Internal Driver also improves but remains lower-scoring.
These results suggest that, despite gains in overall cause reasoning,
recovering the subject's Internal Driver remains harder for most evaluated models.}

\begin{table}[t]
  \centering
  \caption{T5 full-chain reconstruction scores by construct.}
  \label{tab:t5_constructs}
  \scriptsize
  \setlength{\tabcolsep}{3.0pt}
  \renewcommand{\arraystretch}{1.16}
  \begin{tabular*}{\linewidth}{@{\extracolsep{\fill}}lrrrrr@{}}
    \toprule
    & \textbf{Cond.} & \textbf{State} & \textbf{Manif.} & \textbf{Reg.} & \textbf{Effect} \\
    \midrule
    Emotion-Qwen\textsuperscript{*} & 14.09 & 45.69 & 23.20 & 12.09 & 11.18 \\
    \addlinespace[1.2pt]
    LLaVA-OV-70B & 33.49 & 64.10 & 39.28 & 32.08 & 20.39 \\
    Qwen3-VL-32B & 51.59 & 69.61 & 55.83 & 40.64 & 38.21 \\
    \addlinespace[1.2pt]
    GPT-5 (non-thinking) & 66.91 & 70.34 & 62.01 & \textbf{42.55} & 43.80 \\
    Gemini-3-Pro & 63.67 & \textbf{73.22} & 58.72 & 21.46 & 43.25 \\
    GPT-5 (thinking) & \underline{70.29} & 70.54 & \textbf{63.14} & 41.09 & \underline{47.74} \\
    \addlinespace[1.2pt]
    \method{} & \textbf{72.18} & \underline{71.88} & \underline{63.08} & \underline{42.18} & \textbf{53.84} \\
    \bottomrule
  \end{tabular*}
\end{table}

\Needspace{4\baselineskip}
\movetitle{Task~4. Affective Effect Reasoning}
Physical Effect is the lowest-scoring channel for the general-purpose
baselines shown in Table~\ref{tab:t4_fields}, in both the full evaluation and the
other-directed subset (Fig.~\ref{fig:t4_channels}). The relative performance
on Mental and Affective Effect, however, varies across models. In
Fig.~\ref{fig:t4_channels}, closed-source models and Qwen3-VL score higher
on Mental than on Affective Effect, whereas the remaining open-source and
affect-specialized models show the reverse pattern. \method{} achieves
the highest Mental and Physical Effect scores in both evaluation scopes,
while its Affective Effect score remains close to GPT-5 (thinking). Its
gains are therefore uneven across channels, with the largest improvement
over GPT-5 (thinking) on Physical Effect, from \expnum{46.9} to
\expnum{58.0} in Table~\ref{tab:t4_fields}. This improvement also accords
with the T5 diagnostic results in Section~\ref{sec:exp:bottlenecks}, which
show fewer unsupported Effect statements under \method{}.

\Needspace{3\baselineskip}
\movetitle{Task~5. Full Chain Reconstruction}
{Full-chain performance varies markedly by component
(Table~\ref{tab:t5_constructs}). Affective State scores highest across all
baselines, but the lowest-scoring component varies:}
Effect scores lowest for Emotion-Qwen, LLaVA-OneVision-70B, and Qwen3-VL-32B,
while Regulation
scores lowest for the closed-source general-purpose models. For GPT-5,
enabling thinking mainly improves Condition and Effect, with only minor
changes in the remaining components. \method{} further improves on GPT-5
(thinking), raising the overall full-chain score from \expnum{60.79} to
\expnum{63.59}. Its largest component-level gain is on Effect, which
increases from \expnum{47.74} to \expnum{53.84}, while State, Manifestation,
and Regulation change only slightly. These results suggest that
\method{}'s improvement in full-chain reconstruction is particularly
pronounced in identifying downstream consequences.

\begin{figure*}[!t]
  \centering
  \includegraphics[width=\textwidth]{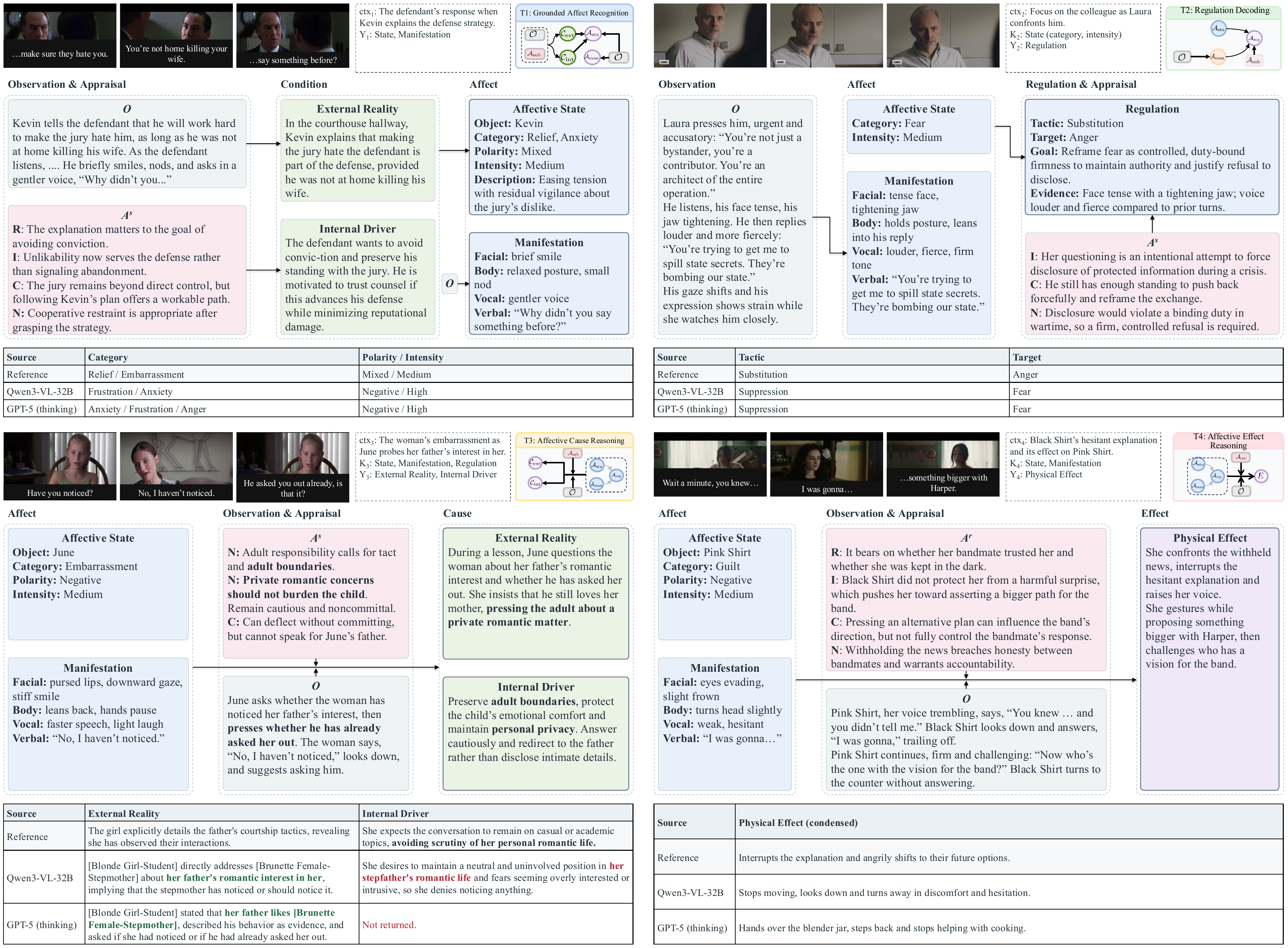}
  \caption{Case studies for T1--T4, with T1 and T2 in the top row and T3 and T4 in the bottom row. Panels show task inputs, selected observations and appraisal readings, construct outputs, and reference and baseline comparisons. Text is condensed from the source records. The T1 example comes from a separate illustrative run. The layouts follow the corresponding task-specific reasoning topologies.}
  \label{fig:case_study}
\end{figure*}

\subsection{Diagnostic Evaluation of Failure Modes}
\label{sec:exp:bottlenecks}

\looseness=-1
These experiments diagnose factual drift through fabricated observable
content in full-chain reconstruction, and surface reading through confusion
between Affective State and Manifestation. Both use judgments blind to
configuration~identity.

\subsubsection{Factual Drift in Full-Chain Reconstruction}
\label{sec:exp:bottleneck_fact}
\leavevmode\par

\exphead{Setup and metric}For T5 outputs, we check factual
statements in External Reality, Manifestation, and Physical Effect against
video frames and subtitles. The fabrication rate is the percentage of
fabricated statements among all judged statements, including unverifiable
ones. Empty fields are tracked separately. Appendix~A.6.1
details the sample, agentic audit, human verification, and results.

\exphead{Results and analysis}Baseline fabrication is most pronounced in
downstream Effect statements (Fig.~\ref{fig:diag_fabrication}).
\method{} reduces the aggregate rate from \expnum{11.0\%} to
\expnum{4.7\%}, with the largest reduction in Effect, from
\expnum{17.3\%} to \expnum{3.3\%}. Supplying observation records alone
also reduces fabrication, showing the contribution of explicit evidence.
The full method substantially reduces unsupported observable content,
particularly in the inferred aftermath of the episode.

\subsubsection{Surface Reading of Affective Manifestations}
\label{sec:exp:bottleneck_surface}
\leavevmode\par

\begin{figure*}[!t]
\begin{minipage}[t]{\columnwidth}
  \vspace{0pt}
  \centering
  \setlength{\abovecaptionskip}{5pt}
  \includegraphics[height=190.8bp]{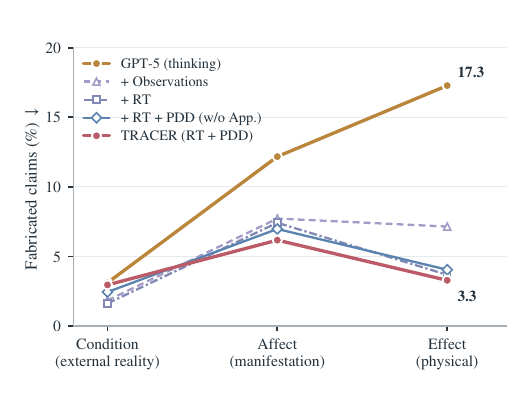}
  \caption{Fabrication rates by chain stage. RT and PDD correspond to
  the topology and derivation modules in Section~\ref{sec:method:derivation}.
  RT includes supplied appraisals, removed in the w/o App. configuration.}
  \label{fig:diag_fabrication}
\end{minipage}\hfill
\begin{minipage}[t]{\columnwidth}
  \vspace{0pt}
  \centering
  \setlength{\abovecaptionskip}{5pt}
  \includegraphics[height=190.8bp]{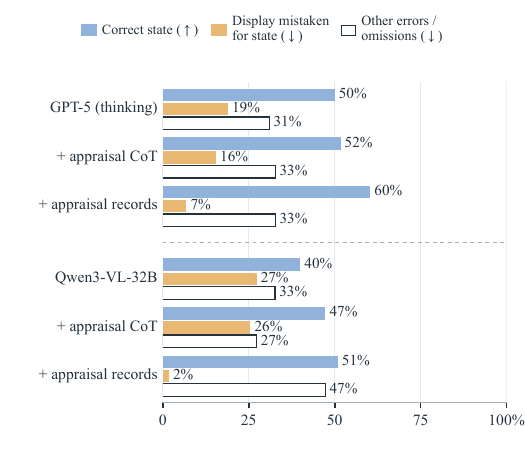}
  \caption{Appraisal interventions for State attribution. Display mistaken
  for state means identifying the outward display as the felt state when
  they differ. All settings share \expnum{58} moments for GPT-5
  and \expnum{55} for Qwen3-VL.}
  \label{fig:diag_surface}
\end{minipage}
\end{figure*}

\exphead{Setup and metric}We test whether appraisal helps models identify
the subject's felt state when it differs from the outward display.
The baseline describes the clip from video frames and subtitles without
being alerted to this mismatch. Appraisal CoT asks the model to analyze
each character's situation along the four appraisal dimensions before
writing the description. In the appraisal-records setting, readings
produced by a separate analysis are supplied alongside the original inputs.
Fig.~\ref{fig:diag_surface} reports the intervention comparisons for
GPT-5 and Qwen3-VL, using the same moments across settings within each
model. We score State attribution at moments where State and Manifestation
differ in category as correct State, Manifestation mistaken for State,
or other errors or omissions. Appendix~A.6.2 details the prompts,
text-based judging, human verification, and control results.

\exphead{Results and analysis}Both models frequently mistake Manifestation
for State in their baseline descriptions (Fig.~\ref{fig:diag_surface}).
Supplying appraisal records improves correct State attribution from
\expnum{50\%} to \expnum{60\%} for GPT-5 and from \expnum{40\%} to
\expnum{51\%} for Qwen3-VL. For GPT-5, appraisal CoT offers little
improvement, whereas supplied records yield more correct State attributions
with the same share of other errors or omissions. For Qwen3-VL, appraisal
CoT raises correct State attribution to \expnum{47\%}, but surface reading
remains at \expnum{26\%}. Supplied records reduce surface reading to
\expnum{2\%}, while other errors or omissions increase to \expnum{47\%}.
Avoiding the displayed
category therefore does not always mean identifying the felt state
correctly. These results support appraisal records for improving State
attribution, although other errors remain.

\subsection{Component Ablation}
\label{sec:exp:trace_tree}
\label{sec:exp:ablation}
\label{sec:exp:method_comparison}
\label{sec:exp:bottleneck_structure}

\begin{table*}[!t]
  \centering
  \caption{Component ablation. Filled circles indicate enabled components
  (open Think: minimal reasoning). RT and PDD denote task-conditioned
  reasoning topology and premise-declared derivation, respectively.
  T2 uses the revised regulation task, as in
  Table~\ref{tab:main_results}.}
  \label{tab:ablation}
  \footnotesize
  \setlength{\tabcolsep}{3pt}
  \renewcommand{\arraystretch}{1.16}
  \begin{tabular*}{\textwidth}{@{\extracolsep{\fill}}l cccccccc rrrrrr@{}}
    \toprule
    & \textbf{Think} & \textbf{Frames} & \textbf{Subs} & \textbf{Obs}
    & \textbf{App} & \textbf{CoT} & \textbf{RT} & \textbf{PDD}
    & \textbf{T1} & \textbf{T2}
    & \textbf{T3} & \textbf{T4} & \textbf{T5} & \textbf{Avg.} \\
    \midrule
    GPT-5 (non-thinking) & $\circ$ & \ding{108} & \ding{108} & $\circ$ & $\circ$ & $\circ$ & $\circ$ & $\circ$ & \underline{56.92} & 39.53 & 70.84 & 55.00 & 57.62 & 55.98 \\
    GPT-5 (thinking) & \ding{108} & \ding{108} & \ding{108} & $\circ$ & $\circ$ & $\circ$ & $\circ$ & $\circ$ & 56.84 & 43.49 & 75.39 & 57.73 & 60.89 & 58.87 \\
    ~~w/ Appraisal-based CoT & \ding{108} & \ding{108} & \ding{108} & $\circ$ & $\circ$ & \ding{108} & $\circ$ & $\circ$ & 55.62 & 47.39 & 80.01 & \underline{58.86} & 61.15 & \underline{60.61} \\
    ~~w/ Observations & \ding{108} & $\circ$ & $\circ$ & \ding{108} & $\circ$ & $\circ$ & $\circ$ & $\circ$ & 50.63 & 46.84 & 77.52 & 47.07 & 59.02 & 56.22 \\
    ~~w/ Observations \& Appraisals & \ding{108} & $\circ$ & $\circ$ & \ding{108} & \ding{108} & $\circ$ & $\circ$ & $\circ$ & 52.56 & 48.37 & 79.73 & 54.20 & \underline{62.83} & 59.54 \\
    ~~w/ RT & \ding{108} & $\circ$ & $\circ$ & \ding{108} & \ding{108} & $\circ$ & \ding{108} & $\circ$ & 53.77 & \underline{51.12} & \underline{80.38} & 55.45 & 61.42 & 60.43 \\
    ~~w/ RT & $\circ$ & $\circ$ & $\circ$ & \ding{108} & \ding{108} & $\circ$ & \ding{108} & $\circ$ & 56.52 & 49.76 & 74.97 & 53.39 & 59.68 & 58.86 \\
    ~~w/ RT + PDD (\method{}) & $\circ$ & $\circ$ & $\circ$ & \ding{108} & \ding{108} & $\circ$ & \ding{108} & \ding{108} & \textbf{61.71} & \textbf{52.12} & \textbf{83.76} & \textbf{59.93} & \textbf{63.39} & \textbf{64.18} \\
    \bottomrule
  \end{tabular*}
\end{table*}

\exphead{Setup}{We assess the contributions of evidence provision,
appraisal-based reasoning, and structured execution on the same fixed list
of 150 items per task (stratified sampling in Appendix~A.5).
These sets differ in size from the main evaluation.}
\textbf{Think} denotes test-time reasoning, with minimal effort used in
\method{}. \textbf{Frames} and \textbf{Subs} are the raw video channels.
\textbf{Obs} is the observation inventory used in their place.
\textbf{App} supplies appraisal records, whereas \textbf{CoT} prompts the
model to generate appraisal-based reasoning within its answer call.
\textbf{RT} supplies the reasoning topology of
Section~\ref{sec:method:derivation} as guidance on construct dependencies
and the use of observation and appraisal premises.
\textbf{PDD} implements that topology by generating conclusion nodes in
dependency order, with explicit premise references and inferred field values.
Configurations using Obs and App share the corresponding records with
\method{}. All configurations are evaluated under the same judge.

\exphead{Effect of supplied material}Replacing raw video input with
observation records improves T2 and T3 but lowers T1 and T4, with losses
in intensity, facial expression, and the recipient's affective reaction.
Adding appraisal records raises every task over observations alone,
most on T4 (\expnum{7.1} points) and T5 (\expnum{3.8}), and lifts the
average above GPT-5 (thinking). Observation records alone therefore
produce task-dependent gains and losses, while adding appraisals yields
a net improvement over the raw-input baseline.

\exphead{Effect of appraisal reasoning}We test whether prompting GPT-5 to
reason through cognitive appraisals within its answer call is sufficient to
improve affective understanding. Appraisal-based CoT raises the average score
from \expnum{58.87} to \expnum{60.61}, but the gains are task-dependent:
T2--T4 improve, T1 declines, and T5 remains nearly unchanged. It still trails
RT + PDD by \expnum{3.6} points, suggesting that appraisal prompting alone
does not recover the full method's gains.

\exphead{Effect of RT}We examine whether providing the task-conditioned
dependency structure as guidance improves reasoning. With the same observation
and appraisal records, adding RT yields a modest average gain of
\expnum{0.89} points and does not improve all tasks. Performance drops further
when reasoning effort is reduced to minimal. These results suggest that
supplying the dependency structure as guidance alone offers limited benefits.

\exphead{Effect of PDD}PDD turns the topology into an explicit execution
procedure. Holding the observation records, appraisal records, RT, and
reasoning effort fixed, adding PDD improves all five tasks and raises the
average score from \expnum{58.86} to \expnum{64.18}. This comparison supports
explicitly executing the topology, with intermediate constructs derived in
dependency order and carried forward as premises, rather than merely
presenting it as guidance.

\subsection{Limitations and Future Work}
\label{sec:exp:limitations}

\benchmark{} uses subject-centered affective episodes as its annotation and
evaluation unit, linking each episode's causes, affect, and consequences
within the surrounding scene. Although a video may contain several annotated
episodes, the current tasks do not explicitly evaluate dependencies between
successive episodes or changes in regulation strategy across them.
Future work can extend the annotations and tasks to examine how earlier
affective experiences shape later appraisals and regulation over longer
interactions.

{At the method level, \method{} reasons over a fixed observation
record extracted before derivation. This representation improves evidence
accessibility, but it can omit fine-grained cues that later become important
for a particular construct. Our ablations reflect this trade-off: replacing
raw inputs with observation records improves some tasks while reducing
performance on affect intensity, facial expression, and the recipient's
affective response. A promising extension is to make evidence acquisition
adaptive, allowing the reasoning process to revisit relevant video intervals
when the current record is insufficient to resolve a construct.
Future work could also reduce the computational overhead of observation
extraction, appraisal, and derivation while preserving evidence grounding.}

\section{Conclusion}
\label{sec:conclusion}

We presented \traceframework{}, a framework for studying affective understanding
as a cognition-shaped process in complex social video. The \blueprint{}
formalizes a subject-centered affective episode through \emph{Condition, Affect,
and Effect}, and \benchmark{} operationalizes this formulation through five
tasks spanning single-stage analysis, cross-stage reasoning, and full-chain
reconstruction. Evaluation across affect-specialized, open-source
general-purpose, and closed-source general-purpose MLLMs shows that current
performance remains substantially below human performance, and that stronger
emotion recognition does not reliably extend to understanding regulation,
causes, effects, or complete affective chains. Error analyses further expose
two persistent weaknesses: surface reading of outward behavior and factual
drift during chain reconstruction.
To provide a baseline for this setting, we developed \method{}, which combines
cognition-grounded appraisal with premise-declared derivation. \method{}
improves performance across all five tasks, with the largest gains in regulation
decoding and cause reasoning, and substantially reduces unsupported content in
full-chain reconstruction. These results suggest that, as multimodal models
become increasingly capable of recognizing affective signals, the cognitive
processes underlying affect deserve greater attention in both modeling and
evaluation.

\bibliographystyle{IEEEtran}
\bibliography{references}

\end{document}


\allowdisplaybreaks[1]
\raggedbottom
\clubpenalty=10000
\widowpenalty=10000
\displaywidowpenalty=10000
\title{Appendix}
\author{}
\markboth{Li \MakeLowercase{\textit{et al.}}: TRACE Appendix}%
{Li \MakeLowercase{\textit{et al.}}: TRACE Appendix}
\twocolumn[{\centering\Huge\sffamily Appendix\par\vspace{10pt}}]
\renewcommand{\thefigure}{S\arabic{figure}}
\renewcommand{\thetable}{S\arabic{table}}
\renewcommand{\theequation}{S\arabic{equation}}
\makeatletter
\setlength{\@fptop}{0pt}
\setlength{\@dblfptop}{0pt}
\setlength{\@fpsep}{10pt}
\setlength{\@dblfpsep}{10pt}
\makeatother

\looseness=-1
\noindent This appendix provides implementation and experimental details
(Appendix~\ref{app:exp_details}), the scoring protocol
(Appendix~\ref{app:scoring}), and details of data construction and annotation
(Appendix~\ref{app:data_quality}). Figures, tables, and equations here carry
the prefix ``S''.

\begin{table*}[t]
  \centering
  \caption{T1 affect recognition. State and Manifestation totals are computed within each item.}
  \label{tab:app_full_t1}
  \footnotesize
  \setlength{\tabcolsep}{3.0pt}
  \renewcommand{\arraystretch}{1.08}
  \begin{tabular*}{\textwidth}{@{\extracolsep{\fill}}lrrrrrrrrrrr@{}}
    \toprule
    & \multicolumn{6}{c}{\textbf{Affective State}} & \multicolumn{5}{c}{\textbf{Manifestation}} \\
    \cmidrule(lr){2-7} \cmidrule(lr){8-12}
    \textbf{Model} & \textbf{Cat.} & \textbf{Desc.} & \textbf{Pol.} & \textbf{Int.} & \textbf{Obj.} & \textbf{All} & \textbf{Facial} & \textbf{Body} & \textbf{Verbal} & \textbf{Vocal} & \textbf{All} \\
    \midrule
    \multicolumn{12}{@{}l}{\textbf{\textit{Affect-specialized}}} \\
    Emotion-LLaMA & \expnum{1.4} & \expnum{0.1} & \expnum{1.1} & \expnum{1.4} & \expnum{0.0} & \expnum{0.8} & \expnum{0.1} & \expnum{0.1} & \expnum{0.1} & \expnum{0.1} & \expnum{0.1} \\
    AffectGPT & \expnum{21.8} & \expnum{23.5} & \expnum{71.5} & \expnum{50.2} & \expnum{56.4} & \expnum{44.8} & \expnum{27.7} & \expnum{22.8} & \expnum{15.6} & \expnum{31.8} & \expnum{24.4} \\
    Emotion-Qwen & \expnum{27.2}\textsuperscript{*} & \expnum{11.7} & \expnum{57.5} & \expnum{35.4} & \expnum{51.8} & \expnum{36.8} & \expnum{30.0} & \expnum{20.4} & \expnum{16.0} & \expnum{23.3} & \expnum{22.1} \\
    \midrule
    \multicolumn{12}{@{}l}{\textbf{\textit{Open-source general-purpose}}} \\
    LLaVA-OneVision-7B & \expnum{31.6} & \expnum{16.3} & \expnum{74.4} & \expnum{51.8} & \expnum{72.8} & \expnum{49.5} & \expnum{27.9} & \expnum{19.1} & \expnum{14.2} & \expnum{21.5} & \expnum{20.7} \\
    LLaVA-NeXT-Video-32B & \expnum{27.6} & \expnum{12.3} & \expnum{79.5} & \expnum{44.3} & \expnum{35.2} & \expnum{39.9} & \expnum{15.0} & \expnum{10.0} & \expnum{10.3} & \expnum{12.1} & \expnum{11.9} \\
    MiniCPM-V-4.5 & \expnum{37.2} & \expnum{27.2} & \expnum{65.7} & \expnum{48.9} & \expnum{60.8} & \expnum{48.0} & \expnum{28.5} & \expnum{25.1} & \expnum{27.0} & \expnum{29.1} & \expnum{26.6} \\
    GLM-4.6V-Flash-9B & \expnum{29.4} & \expnum{35.9} & \expnum{77.0} & \expnum{57.2} & \expnum{63.0} & \expnum{52.6} & \expnum{34.3} & \expnum{20.4} & \expnum{34.8} & \expnum{28.7} & \expnum{29.0} \\
    LLaVA-OneVision-70B & \expnum{40.1} & \expnum{36.8} & \expnum{\underline{80.1}} & \expnum{61.1} & \expnum{67.7} & \expnum{57.3} & \expnum{32.9} & \expnum{23.1} & \expnum{31.0} & \expnum{27.5} & \expnum{28.2} \\
    Qwen3-Omni-30B & \expnum{42.5} & \expnum{43.6} & \expnum{76.1} & \expnum{59.2} & \expnum{52.5} & \expnum{54.9} & \expnum{42.0} & \expnum{32.1} & \expnum{40.4} & \expnum{38.9} & \expnum{37.7} \\
    InternVL3.5-38B & \expnum{\underline{44.0}} & \expnum{37.0} & \expnum{77.3} & \expnum{61.4} & \expnum{\underline{74.7}} & \expnum{59.0} & \expnum{40.0} & \expnum{27.5} & \expnum{30.3} & \expnum{31.0} & \expnum{31.7} \\
    Qwen3-VL-32B & \expnum{38.8} & \expnum{48.1} & \expnum{\textbf{81.4}} & \expnum{64.2} & \expnum{67.8} & \expnum{60.1} & \expnum{47.2} & \expnum{36.6} & \expnum{44.8} & \expnum{43.0} & \expnum{41.8} \\
    \midrule
    \multicolumn{12}{@{}l}{\textbf{\textit{Closed-source general-purpose}}} \\
    GPT-5 (non-thinking) & \expnum{36.1} & \expnum{55.6} & \expnum{69.1} & \expnum{\underline{64.5}} & \expnum{73.0} & \expnum{59.7} & \expnum{56.9} & \expnum{46.2} & \expnum{60.1} & \expnum{52.4} & \expnum{52.6} \\
    Gemini-3-Pro & \expnum{\textbf{48.1}} & \expnum{53.7} & \expnum{\textbf{81.4}} & \expnum{58.2} & \expnum{\underline{74.7}} & \expnum{\underline{63.3}} & \expnum{48.5} & \expnum{37.3} & \expnum{55.4} & \expnum{45.6} & \expnum{45.4} \\
    GPT-5 (thinking) & \expnum{41.5} & \expnum{\underline{57.3}} & \expnum{71.3} & \expnum{63.3} & \expnum{62.5} & \expnum{59.2} & \expnum{\textbf{59.2}} & \expnum{\underline{47.2}} & \expnum{\underline{61.6}} & \expnum{\underline{54.2}} & \expnum{\underline{54.0}} \\
    \midrule
    \multicolumn{12}{@{}l}{\textbf{\textit{Ours}}} \\
    \method{} & \expnum{40.5} & \expnum{\textbf{60.0}} & \expnum{77.5} & \expnum{\textbf{66.5}} & \expnum{\textbf{79.9}} & \expnum{\textbf{64.9}} & \expnum{\underline{57.7}} & \expnum{\textbf{55.6}} & \expnum{\textbf{70.6}} & \expnum{\textbf{56.6}} & \expnum{\textbf{58.4}} \\
    \bottomrule
  \end{tabular*}

\end{table*}

\begin{table*}[t]
  \centering
  \caption{Field-level results for T2 regulation decoding, T3 cause reasoning, and T4 effect reasoning.}
  \label{tab:app_full_t234}
  \label{tab:app_full_t2}
  \label{tab:app_full_t3}
  \label{tab:app_full_t4}
  \footnotesize
  \setlength{\tabcolsep}{3.0pt}
  \renewcommand{\arraystretch}{1.08}
  \begin{tabular*}{\textwidth}{@{\extracolsep{\fill}}lrrrrrrrrr@{}}
    \toprule
    & \multicolumn{4}{c}{\textbf{T2 Regulation}} & \multicolumn{2}{c}{\textbf{T3 Cause}} & \multicolumn{3}{c}{\textbf{T4 Effect}} \\
    \cmidrule(lr){2-5} \cmidrule(lr){6-7} \cmidrule(lr){8-10}
    \textbf{Model} & \textbf{Tactic} & \textbf{Target} & \textbf{Goal} & \textbf{Evidence} & \textbf{External} & \textbf{Internal} & \textbf{Mental} & \textbf{Affective} & \textbf{Physical} \\
    \midrule
    \multicolumn{10}{@{}l}{\textbf{\textit{Affect-specialized}}} \\
    Emotion-LLaMA & \expnum{10.2}\textsuperscript{*} & \expnum{5.6} & \expnum{4.3} & \expnum{1.5} & \expnum{2.5}\textsuperscript{\dag} & \expnum{0.2} & \expnum{0.0}\textsuperscript{\dag} & \expnum{0.6} & \expnum{0.2} \\
    AffectGPT & \expnum{7.0}\textsuperscript{\dag} & \expnum{1.9} & \expnum{6.2} & \expnum{1.6} & \expnum{19.6}\textsuperscript{*} & \expnum{10.8} & \expnum{12.4}\textsuperscript{*} & \expnum{22.1} & \expnum{12.6} \\
    Emotion-Qwen & \expnum{36.7}\textsuperscript{*} & \expnum{28.0} & \expnum{13.5} & \expnum{10.0} & \expnum{24.4}\textsuperscript{*} & \expnum{9.1} & \expnum{2.1} & \expnum{19.6} & \expnum{15.5} \\
    \midrule
    \multicolumn{10}{@{}l}{\textbf{\textit{Open-source general-purpose}}} \\
    LLaVA-OneVision-7B & \expnum{45.4} & \expnum{15.9} & \expnum{23.2} & \expnum{5.2} & \expnum{26.9} & \expnum{17.9} & \expnum{2.3} & \expnum{6.9} & \expnum{2.9} \\
    LLaVA-NeXT-Video-32B & \expnum{47.9} & \expnum{8.4} & \expnum{28.9} & \expnum{4.1} & \expnum{28.9}\textsuperscript{*} & \expnum{15.3} & \expnum{23.1} & \expnum{33.9} & \expnum{15.2} \\
    MiniCPM-V-4.5 & \expnum{40.9} & \expnum{32.7} & \expnum{40.1} & \expnum{13.0} & \expnum{43.8} & \expnum{36.4} & \expnum{32.5} & \expnum{37.4} & \expnum{17.9} \\
    GLM-4.6V-Flash-9B & \expnum{44.6} & \expnum{5.6} & \expnum{36.9} & \expnum{6.4} & \expnum{43.5} & \expnum{37.2} & \expnum{41.8} & \expnum{43.4} & \expnum{8.2} \\
    LLaVA-OneVision-70B & \expnum{46.9} & \expnum{5.6} & \expnum{30.1} & \expnum{8.7} & \expnum{49.1} & \expnum{37.0} & \expnum{38.1} & \expnum{40.7} & \expnum{23.3} \\
    Qwen3-Omni-30B & \expnum{47.4} & \expnum{22.4} & \expnum{46.5} & \expnum{17.6} & \expnum{52.3} & \expnum{47.9} & \expnum{47.4} & \expnum{47.4} & \expnum{28.6} \\
    InternVL3.5-38B & \expnum{47.4} & \expnum{5.6} & \expnum{30.3} & \expnum{8.5} & \expnum{47.5} & \expnum{39.9} & \expnum{36.4} & \expnum{41.7} & \expnum{21.5} \\
    Qwen3-VL-32B & \expnum{48.9} & \expnum{26.2} & \expnum{51.1} & \expnum{18.1} & \expnum{60.8} & \expnum{57.9} & \expnum{53.5} & \expnum{49.0} & \expnum{34.6} \\
    \midrule
    \multicolumn{10}{@{}l}{\textbf{\textit{Closed-source general-purpose}}} \\
    GPT-5 (non-thinking) & \expnum{47.4} & \expnum{28.0} & \expnum{54.8} & \expnum{\underline{23.6}} & \expnum{72.9} & \expnum{62.6} & \expnum{59.4} & \expnum{\textbf{57.8}} & \expnum{43.3} \\
    Gemini-3-Pro & \expnum{49.5} & \expnum{31.6} & \expnum{51.2} & \expnum{18.1} & \expnum{71.6} & \expnum{65.8} & \expnum{58.8} & \expnum{52.6} & \expnum{45.9} \\
    GPT-5 (thinking) & \expnum{\underline{50.9}} & \expnum{\textbf{46.7}} & \expnum{\underline{56.2}} & \expnum{\underline{23.6}} & \expnum{\underline{78.3}} & \expnum{\underline{72.0}} & \expnum{\underline{60.5}} & \expnum{56.8} & \expnum{\underline{46.9}} \\
    \midrule
    \multicolumn{10}{@{}l}{\textbf{\textit{Ours}}} \\
    \method{} & \expnum{\textbf{53.8}} & \expnum{\underline{37.7}} & \expnum{\textbf{66.5}} & \expnum{\textbf{50.3}} & \expnum{\textbf{87.1}} & \expnum{\textbf{76.4}} & \expnum{\textbf{64.5}} & \expnum{\underline{57.0}} & \expnum{\textbf{58.0}} \\
    \bottomrule
  \end{tabular*}

\end{table*}

\appendices

\section{Implementation and Experimental Details}
\label{app:exp_details}

\subsection{Inference Settings}
\label{app:baseline_configs}

During inference, we provide all evaluated methods with the same task
instructions and subtitle transcripts. Alongside these textual inputs,
the models receive frames sampled across the complete clip, with the
model-specific budgets listed below. Audio is provided only to Qwen3-Omni-30B.
We retain the released inference
settings for affect-specialized models. GPT-5 uses the provider's default
reasoning setting for the thinking baseline and minimal reasoning effort
for the non-thinking baseline.
\begin{center}
\small
\begin{tabularx}{\columnwidth}{@{}>{\raggedright\arraybackslash}Xl@{}}
\toprule
\textbf{Model} & \textbf{Video budget} \\
\midrule
\multicolumn{2}{@{}l}{\textbf{Open-source general-purpose}}\\
\mbox{LLaVA-OneVision-7B/70B}, \mbox{LLaVA-NeXT-Video-32B} & 32 frames \\
\mbox{InternVL3.5-38B}, \mbox{Qwen3-Omni-30B}, \mbox{Qwen3-VL-32B} & 60 frames \\
GLM-4.6V-Flash-9B & 8 frames \\
MiniCPM-V-4.5 & 5 fps \\
\midrule
\multicolumn{2}{@{}l}{\textbf{Closed-source general-purpose}}\\
GPT-5 (thinking/non-thinking), \mbox{Gemini-3-Pro} & 8 frames \\
\bottomrule
\end{tabularx}
\end{center}
\runin{TRACER settings.}%
We use Qwen3-VL-32B to extract one observation record per video and reuse
that record across tasks, ablations, and diagnostics, following the
procedure in Appendix~\ref{app:observation}. We then use GPT-5 for appraisal,
anchor alignment, and structured derivation, setting its reasoning effort
to low for appraisal and minimal for the latter two operations. For T5,
we derive Condition, Affect, and Effect through separate calls.

\begin{table}[H]
  \centering
  \caption{Answer-generation calls and token usage, averaged over five tasks.}
  \label{tab:app_token_cost}
  \footnotesize
  \setlength{\tabcolsep}{3pt}
  \begin{tabularx}{\columnwidth}{@{}>{\raggedright\arraybackslash}Xrrrr@{}}
    \toprule
    \textbf{Configuration} & \textbf{Calls} & \textbf{Input} & \textbf{Output} & \shortstack{\textbf{Hidden}\\\textbf{reasoning}} \\
    \midrule
    GPT-5 (thinking) & \expnum{1} & \expnum{1.7k} & \expnum{1.25k} & \expnum{1.06k} \\
    GPT-5 (thinking) + Obs & \expnum{1} & \expnum{2.2k} & \expnum{1.10k} & \expnum{0.90k} \\
    GPT-5 (thinking) + Obs + App & \expnum{1} & \expnum{2.9k} & \expnum{1.13k} & \expnum{0.93k} \\
    \method{} & \expnum{1.4} & \expnum{17.3k} & \expnum{0.71k} & \expnum{0} \\
    \bottomrule
  \end{tabularx}
\end{table}

\runin{Token use.}%
We measure token use on five questions per task and average equally across
the five tasks (Table~\ref{tab:app_token_cost}). T1--T4 each use one
derivation call, while T5 uses three, giving \method{} an average of 1.4
calls per question. Output counts include provider-reported hidden
reasoning tokens. The table reports answer generation; observation records
are generated once per video on a local GPU, and appraisal uses 9.2k input
and 1.0k output tokens per question. These records are reused across
configurations. Each \method{} derivation call includes the method
specification and observation record, resulting in higher
input token use.

\begin{figure}[!t]
  \centering
  \input{appendix/drafts/appendix_polish_assets/observer_prompt.tex}
  \caption{Observation extraction. (a) Abridged observer prompt with simplified input labels; [\ldots] marks omissions and braces denote inserted inputs. (b) Output format. (c) One observation of a record.}
  \label{fig:app_observer_prompt}
\end{figure}

\subsection{Observation Extraction}
\label{app:observation}

To construct the observation record, we sample each video at 1~fps, up to
256 frames, and provide these frames to Qwen3-VL-32B-Instruct with
timestamped subtitles and model-generated speaker/action attribution,
using a character roster to align person references.
The character roster is compiled from the subject, object, and recipient
identifiers provided in the QA questions. Speaker/action attribution is
generated by a preceding model stage using the full dialogue.
The observation call does not include the full task question.
We ask the model to describe 8 to 12 moments depending on clip length,
allowing up to 8{,}192 output tokens. For each moment, it records an
identifier, time span, characters, and a description grounded in the frames
and subtitles. The prompt and output example in Fig.~\ref{fig:app_observer_prompt}
show how these records are organized, with emotion labels, causal
statements, and selection by relevance excluded.

\subsection{Appraisal Prompt}
\label{app:appraisal_prompt}

For each appraisal request, we provide the observation record and specify
the character and the group of items to interpret. The model returns one
evaluative proposition for each of the four dimensions. During subsequent
derivation, it selects which of these propositions to cite as premises.
Fig.~\ref{fig:app_appraisal_prompt} reproduces the core instructions,
omitting the surrounding method description and item-specific input.

\begin{figure}[!t]
  \centering
  \input{appendix/drafts/appendix_polish_assets/appraisal_prompt.tex}
  \caption{Core appraisal instructions. Wording is retained from the recorded prompt, with formatting adjusted.}
  \label{fig:app_appraisal_prompt}
\end{figure}

\subsection{Field-Level Results}
\label{app:within_task_breakdown}

\begin{table}[!t]
  \centering
  \caption{T5 full-chain reconstruction. Full is the overall T5 score.}
  \label{tab:app_full_t5}
  \footnotesize
  \setlength{\tabcolsep}{2.5pt}
  \renewcommand{\arraystretch}{1.08}
  \begin{tabular*}{\columnwidth}{@{\extracolsep{\fill}}lrrrrrr@{}}
    \toprule
    \textbf{Model} & \textbf{Cond.} & \textbf{State} & \textbf{Manif.} & \textbf{Reg.} & \textbf{Effect} & \textbf{Full} \\
    \midrule
    \multicolumn{7}{@{}l}{\textbf{\textit{Affect-specialized}}} \\
    Emotion-LLaMA & \expnum{0.00}\textsuperscript{\dag} & \expnum{1.38} & \expnum{0.00} & \expnum{0.00} & \expnum{0.00} & \expnum{0.20} \\
    AffectGPT & \expnum{31.36}\textsuperscript{*} & \expnum{47.28} & \expnum{0.34} & \expnum{1.32} & \expnum{9.89} & \expnum{20.87} \\
    Emotion-Qwen & \expnum{14.09}\textsuperscript{*} & \expnum{45.69} & \expnum{23.20} & \expnum{12.09} & \expnum{11.18} & \expnum{19.13} \\
    \midrule
    \multicolumn{7}{@{}l}{\textbf{\textit{Open-source general-purpose}}} \\
    LLaVA-OneVision-7B & \expnum{1.95}\textsuperscript{\dag} & \expnum{15.86} & \expnum{6.65} & \expnum{3.17} & \expnum{3.99} & \expnum{5.47} \\
    LLaVA-NeXT-Video-32B & \expnum{13.80}\textsuperscript{*} & \expnum{28.58} & \expnum{12.35} & \expnum{13.67} & \expnum{7.71} & \expnum{13.80} \\
    MiniCPM-V-4.5 & \expnum{26.95} & \expnum{58.14} & \expnum{28.59} & \expnum{12.40} & \expnum{21.22} & \expnum{29.43} \\
    GLM-4.6V-Flash-9B & \expnum{33.43} & \expnum{57.82} & \expnum{43.43} & \expnum{15.00} & \expnum{16.57} & \expnum{32.36} \\
    LLaVA-OneVision-70B & \expnum{33.49} & \expnum{64.10} & \expnum{39.28} & \expnum{32.08} & \expnum{20.39} & \expnum{34.62} \\
    Qwen3-Omni-30B & \expnum{43.93} & \expnum{64.17} & \expnum{46.62} & \expnum{33.11} & \expnum{27.76} & \expnum{41.78} \\
    InternVL3.5-38B & \expnum{36.99} & \expnum{64.93} & \expnum{42.47} & \expnum{17.08} & \expnum{24.28} & \expnum{37.07} \\
    Qwen3-VL-32B & \expnum{51.59} & \expnum{69.61} & \expnum{55.83} & \expnum{40.64} & \expnum{38.21} & \expnum{50.16} \\
    \midrule
    \multicolumn{7}{@{}l}{\textbf{\textit{Closed-source general-purpose}}} \\
    GPT-5 (non-thinking) & \expnum{66.91} & \expnum{70.34} & \expnum{62.01} & \expnum{\textbf{42.55}} & \expnum{43.80} & \expnum{58.22} \\
    Gemini-3-Pro & \expnum{63.67} & \expnum{\textbf{73.22}} & \expnum{58.72} & \expnum{21.46} & \expnum{43.25} & \expnum{56.05} \\
    GPT-5 (thinking) & \expnum{\underline{70.29}} & \expnum{70.54} & \expnum{\textbf{63.14}} & \expnum{41.09} & \expnum{\underline{47.74}} & \expnum{\underline{60.79}} \\
    \midrule
    \multicolumn{7}{@{}l}{\textbf{\textit{Ours}}} \\
    \method{} & \expnum{\textbf{72.18}} & \expnum{\underline{71.88}} & \expnum{\underline{63.08}} & \expnum{\underline{42.18}} & \expnum{\textbf{53.84}} & \expnum{\textbf{63.59}} \\
    \bottomrule
  \end{tabular*}

\end{table}

Tables~\ref{tab:app_full_t1}--\ref{tab:app_full_t5} extend the selected
comparisons in Section~6.4 of the main paper to all evaluated models.

We average each field's score over the items that query it, whereas task
scores are aggregated within each item before averaging across items.
An asterisk flags parsing rates below 90\% and a dagger rates below 50\%;
failed outputs score zero in both cases. The parsing rates and the means
over parsed outputs are provided in Appendix~\ref{app:score_reporting}.

\needspace{5\baselineskip}
\subsection{Ablation and Model-Family Comparisons}
\label{app:ablation_pool}

\runin{Sampling.}%
The ablation uses a fixed list of 150 items per task. We divide the
per-item scores of GPT-5 (thinking), on a 0--1 scale, into five equal-width
bands: $[0,0.2)$, $[0.2,0.4)$, $[0.4,0.6)$, $[0.6,0.8)$, and $[0.8,1]$.
For each task, the 150 items are allocated across these bands in proportion
to their shares in the full pool.

\runin{Model-family normalization.}%
\label{app:model_family_preference}
For the three baseline families in Fig.~6b of the main paper,
let $S_{g,t}$ be the mean score of family $g$ on task $t$, excluding
\method{}. The heatmap uses double-centered scores:
\begin{equation}
P_{g,t}=S_{g,t}-\bar S_g-\bar S_t+\bar S,
\label{eq:model_family_preference}
\end{equation}
where $\bar S_g$, $\bar S_t$, and $\bar S$ are the family, task, and grand
means. Positive values indicate relative strengths. The Mean row reports
the uncentered family scores.

\begin{figure*}[t]
  \centering
  \input{appendix/drafts/appendix_polish_assets/audit_rubric.tex}
  \caption{Rubric of the unsupported-content audit. Slots s1--s6 are the audited fields. Labels: S supported, L misplaced, F fabricated, U unverifiable. The off-topic denominator refers to the answer-level off-topic rate, not the statement-level fabrication rate.}
  \label{fig:app_audit_rubric}
\end{figure*}

\begin{figure*}[t]
  \centering
  \input{appendix/drafts/appendix_polish_assets/surface_prompts.tex}
  \caption{Prompts of the surface-reading diagnostic. (a) Description prompt. (b) Block appended in the appraisal-records setting. (c) Instruction prepended in the appraisal-CoT setting.}
  \label{fig:app_surface_prompts}
\end{figure*}

\subsection{Diagnostic Experiments}
\label{app:diagnostic_protocols}

\subsubsection{\normalfont\bfseries Factual Drift}
\label{app:audit_protocol}

\runin{Assessment.}%
To diagnose factual drift, we audit 460 answers to 92 T5 items under the five configurations in
Table~\ref{tab:audit_results}, checking External Reality, the four
Manifestation channels, and Physical Effect against video evidence.
\looseness=-1
For each answer, we provide an agent using \texttt{claude-opus-5} with timestamped subtitles
without speaker labels and a common frame grid. The agent can read files
and extract additional frames as needed, inspecting an average of 23 time
points per answer. Observation records help locate relevant moments and
align speaker references. Events are verified against video frames and
subtitles. We conceal configuration identities with
randomized path codes and withhold reference answers, answers from other
configurations, and audio.

\runin{Decision rule.}%
The agent splits each field into atomic factual statements and assigns
one of four labels using the rubric in Fig.~\ref{fig:app_audit_rubric}:
\begin{center}
\small
\renewcommand{\arraystretch}{1.12}
\begin{tabularx}{\columnwidth}{@{}lX@{}}
\toprule
\textbf{Label} & \textbf{Criterion}\\
\midrule
S: Supported & Evidence confirms the event and character.\\
L: Misplaced & The statement describes another character or moment.\\
F: Fabricated & Relevant frames or subtitles cover the claimed content but do not support it.\\
U: Unverifiable & Off-screen events or missing audio or dialogue prevent verification.\\
\bottomrule
\end{tabularx}
\end{center}
Writing $n_S,n_L,n_F,n_U$ for the label counts, we report
\[
\text{Fabrication rate}=\frac{n_F}{n_S+n_L+n_F+n_U}.
\]
Across the audited answers, 10{,}579 statements enter this denominator.
We treat empty fields as omissions rather than statements and exclude
off-topic answers, which also yield no statements, from the fabrication
rate. The single off-topic answer is reported separately (1 of 460).

\runin{Human verification.}%
Each answer receives one primary agentic assessment. Human reviewers
reassess 64 distinct answers, including 20 with multiple reassessments
in the pilot phase. Pairing each reassessment with its primary assessment
gives 105 pairs and 630 field-slot comparisons. Judgments on whether a slot
contains fabricated content agree in 570 comparisons
(90.5\%, $\kappa=0.59$). These comparisons are not independent, since
some answers contribute multiple pairs.

\begin{table}[!t]
  \centering
  \caption{T5 fabrication rates by stage (\%, lower is better).}
  \label{tab:audit_results}
  \footnotesize
  \setlength{\tabcolsep}{4.5pt}
  \renewcommand{\arraystretch}{1.08}
  \begin{tabularx}{\columnwidth}{@{}Xrrr@{\hspace{9pt}}rrr@{}}
    \toprule
    \textbf{Configuration} & \textbf{Cond.} & \textbf{Affect} & \textbf{Effect}
    & \textbf{All} & \textbf{Unver.} & \textbf{Mispl.} \\
    \midrule
    GPT-5 (thinking) & \expnum{3.1} & \expnum{12.2} & \expnum{17.3} & \expnum{11.0} & \expnum{12.6} & \expnum{9.5} \\
    ~~+ Observations & \expnum{\underline{1.8}} & \expnum{7.7} & \expnum{7.1} & \expnum{6.3} & \expnum{12.6} & \expnum{6.5} \\
    ~~+ RT & \expnum{\textbf{1.6}} & \expnum{7.4} & \expnum{\underline{3.7}} & \expnum{5.5} & \expnum{13.0} & \expnum{5.0} \\
    ~~+ RT + PDD$^{\ddagger}$ & \expnum{2.4} & \expnum{\underline{7.0}} & \expnum{4.0} & \expnum{\underline{5.0}} & \expnum{13.5} & \expnum{\textbf{3.6}} \\
    \midrule
    \method{} & \expnum{3.0} & \expnum{\textbf{6.2}} & \expnum{\textbf{3.3}} & \expnum{\textbf{4.7}} & \expnum{12.9} & \expnum{\underline{3.9}} \\
    \bottomrule
  \end{tabularx}
  \tabnote{Unver.: unverifiable, not ranked. Mispl.: misplaced. $^{\ddagger}$Without appraisal records.}
\end{table}

\runin{Paired comparison.}%
Pairing fabrication rates by item, the GPT-5 (thinking) minus \method{}
difference is 6.3 percentage points
($t=6.69$). The smaller differences from + RT and
+ RT + PDD (w/o App.) are not statistically significant. Unverifiable
rates remain similar across configurations.

\begin{figure*}[t]
  \centering
  \setlength{\fboxsep}{6pt}\setlength{\fboxrule}{0.4pt}%
  \fcolorbox{black!35}{white}{\begin{minipage}{\dimexpr\textwidth-2\fboxsep-2\fboxrule\relax}
  \input{appendix/drafts/appendix_polish_assets/intervention_case.tex}
  \end{minipage}}
  \caption{Appraisal intervention on a regulated moment. Top: frames and dialogue of the annotated episode, with the focal utterance in bold. Bottom: the reference annotation and excerpts of the model descriptions without and with appraisal records. Without the records, both models take the laughter at face value. With them, both identify the annotated State.}
  \label{fig:app_intervention_case}
\end{figure*}

\subsubsection{\normalfont\bfseries Surface Reading}
\label{app:surface_protocol}

\runin{Assessment.}%
We ask models to describe each clip using the prompts in
Fig.~\ref{fig:app_surface_prompts}, without directing them to regulation
or a particular moment. From these descriptions, we assess State
attribution at moments where State and Manifestation differ in category,
using unregulated moments as controls. We give a judge using \texttt{claude-opus-5} the description
and annotated State and Manifestation categories, but no video or subtitles,
and conceal model and configuration identities with randomized~path~codes.

\runin{Decision rule.}%
For each character--moment, the judge assigns one outcome:
\begin{itemize}[leftmargin=1.2em,itemsep=2pt,parsep=0pt,topsep=3pt]
\item \textbf{Correct State}: the description identifies the annotated State,
including a synonym or a main component of a mixed state. Naming both
State and Manifestation also counts as correct.
\item \textbf{Surface reading}: Manifestation is mistaken for State.
This outcome does not apply to unregulated controls.
\item \textbf{Other/omitted}: another error or no attribution.
\end{itemize}

\runin{Human verification.}%
After the Opus judge assigns one classification to each character--moment,
a human reviewer independently re-evaluates a subset drawn from the
comparison and intervention sets. The two judgments agree on 91 of 98
descriptions (92.9\%).

Table~\ref{tab:surface_results} reports regulated moments and unregulated
controls separately. Table~\ref{tab:surface_intervention} extends the
intervention comparison to four models. For the open-source models,
reduced surface reading is accompanied by an increase in other errors
or omissions, so it does not translate fully into correct State~attribution.

\needspace{5\baselineskip}
Fig.~\ref{fig:app_intervention_case} illustrates the role of appraisal in
interpreting the same observable cue. The supplied records connect the
guest's preference for a more worthwhile conversation with the social cost
of criticizing it. Both models then interpret her laughter as softening
disdain rather than expressing amusement.

\begin{table*}[!t]
\begin{minipage}[t]{0.48\textwidth}
  \centering
  \caption{State attribution with and without regulation (\%).}
  \label{tab:surface_results}
  \footnotesize
  \setlength{\tabcolsep}{3pt}
  \renewcommand{\arraystretch}{1.08}
  \begin{tabular*}{\linewidth}{@{\extracolsep{\fill}}lrrrr@{}}
    \toprule
    \textbf{Moment} & \textbf{$n$}
    & \shortstack{\textbf{Correct}\\\textbf{State} $\uparrow$}
    & \shortstack{\textbf{Surface}\\\textbf{reading} $\downarrow$}
    & \shortstack{\textbf{Other/}\\\textbf{omitted}} \\
    \midrule
    \multicolumn{5}{@{}l}{\textbf{\textit{GPT-5 (thinking)}}} \\
    Regulated & \expnum{58} & \expnum{50.0} & \expnum{19.0} & \expnum{31.0} \\
    Unregulated & \expnum{10} & \expnum{50.0} & -- & \expnum{50.0} \\
    \midrule
    \multicolumn{5}{@{}l}{\textbf{\textit{Qwen3-VL-32B}}} \\
    Regulated & \expnum{58} & \expnum{37.9} & \expnum{27.6} & \expnum{34.5} \\
    Unregulated & \expnum{10} & \expnum{90.0} & -- & \expnum{10.0} \\
    \midrule
    \multicolumn{5}{@{}l}{\textbf{\textit{AffectGPT}}} \\
    Regulated & \expnum{52} & \expnum{11.5} & \expnum{1.9} & \expnum{86.5} \\
    Unregulated & \expnum{10} & \expnum{50.0} & -- & \expnum{50.0} \\
    \bottomrule
  \end{tabular*}
  \tabnote{Surface reading: Manifestation mistaken for State.
  Other/omitted: other errors or omissions. --: not applicable.}
\end{minipage}\hfill
\begin{minipage}[t]{0.48\textwidth}
  \centering
  \caption{Effect of appraisal on state attribution (\%).}
  \label{tab:surface_intervention}
  \footnotesize
  \setlength{\tabcolsep}{3pt}
  \renewcommand{\arraystretch}{1.08}
  \begin{tabular*}{\linewidth}{@{\extracolsep{\fill}}lrrrr@{}}
    \toprule
    \textbf{Setting} & \textbf{$n$}
    & \shortstack{\textbf{Correct}\\\textbf{State} $\uparrow$}
    & \shortstack{\textbf{Surface}\\\textbf{reading} $\downarrow$}
    & \shortstack{\textbf{Other/}\\\textbf{omitted}} \\
    \midrule
    \multicolumn{5}{@{}l}{\textbf{\textit{GPT-5 (thinking)}}} \\
    Baseline & \expnum{58} & \expnum{50.0} & \expnum{19.0} & \expnum{31.0} \\
    Appraisal CoT & \expnum{58} & \expnum{51.7} & \expnum{15.5} & \expnum{32.8} \\
    Appraisal records & \expnum{58} & \textbf{\expnum{60.3}} & \textbf{\expnum{6.9}} & \expnum{32.8} \\
    \midrule
    \multicolumn{5}{@{}l}{\textbf{\textit{Qwen3-VL-32B}}} \\
    Baseline & \expnum{55} & \expnum{40.0} & \expnum{27.3} & \expnum{32.7} \\
    Appraisal CoT & \expnum{55} & \expnum{47.3} & \expnum{25.5} & \expnum{27.3} \\
    Appraisal records & \expnum{55} & \textbf{\expnum{50.9}} & \textbf{\expnum{1.8}} & \expnum{47.3} \\
    \midrule
    \multicolumn{5}{@{}l}{\textbf{\textit{InternVL3.5-38B}}} \\
    Baseline & \expnum{55} & \expnum{38.2} & \expnum{21.8} & \expnum{40.0} \\
    Appraisal records & \expnum{55} & \textbf{\expnum{43.6}} & \textbf{\expnum{9.1}} & \expnum{47.3} \\
    \midrule
    \multicolumn{5}{@{}l}{\textbf{\textit{GLM-4.6V-Flash}}} \\
    Baseline & \expnum{55} & \expnum{38.2} & \expnum{18.2} & \expnum{43.6} \\
    Appraisal records & \expnum{55} & \textbf{\expnum{47.3}} & \textbf{\expnum{3.6}} & \expnum{49.1} \\
    \bottomrule
  \end{tabular*}
  \tabnote{Baseline: no appraisal intervention.}
\end{minipage}
\end{table*}

\begin{figure}[t]
  \centering
  \includegraphics[width=\columnwidth]{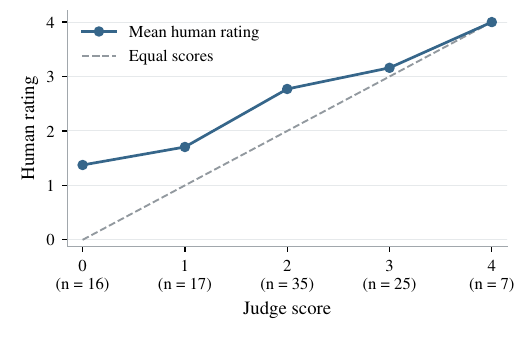}
  \caption{Human--judge calibration on 100 free-text fields. Points show mean human ratings, with group sizes below the axis. The dashed line indicates equal scores.}
  \label{fig:app_judge_trend}
\end{figure}

\subsection{Human Solvability and Judge Calibration}
\label{sec:exp:reliability}
\label{app:validation_sampling}

\runin{Human solvability.}%
We select 50 questions per task from 250 distinct videos using reference
annotations and video metadata, without consulting model predictions or
evaluation results. We stratify by the first annotated affect category
for T1, regulation tactic for T2, whether Internal Driver is queried for T3,
the queried Effect channel (mental, affective, or physical) for T4, and
whether the chain involves regulation for T5. Sampling follows the
eligible pool's stratum proportions.
Strata expected to contribute fewer than two questions are merged before
selection, without additional weighting. Videos used in other paper analyses,
including annotation-agreement assessment and case studies, are excluded.
The seed, stratum proportions, exclusions, and ordered sample list are frozen
before answer collection. Unavailable items are replaced by the next eligible
item on the list, with skips and replacements logged. Two human participants
independently answer each question using the same inputs as the models.
Table~\ref{tab:human_matched} compares their mean scores with model scores
on the same questions. Their task and component means also appear in
Table~2 and Fig.~6a of the main paper.

\begin{table}[!t]
  \centering
  \caption{Human and model performance on the same 50 questions per task.
  Human scores average two participants; Avg. averages the five tasks.}
  \label{tab:human_matched}
  \scriptsize
  \setlength{\tabcolsep}{2pt}
  \renewcommand{\arraystretch}{1.12}
  \begin{tabular*}{\columnwidth}{@{\extracolsep{\fill}}lrrrrrr@{}}
    \toprule
    \textbf{Model / Human} & \textbf{T1} & \textbf{T2} & \textbf{T3} & \textbf{T4} & \textbf{T5} & \textbf{Avg.} \\
    \midrule
    Human & 83.60 & 72.32 & 85.55 & 78.90 & 75.40 & 79.15 \\
    \midrule
    GPT-5 (thinking) & \underline{57.97} & \underline{41.87} & \underline{75.80} & 57.61 & \underline{60.48} & \underline{58.75} \\
    GPT-5 (non-thinking) & 55.21 & 36.00 & 68.12 & \textbf{61.06} & 59.09 & 55.89 \\
    Gemini-3-Pro & 55.59 & 41.53 & 68.00 & 60.19 & 54.18 & 55.90 \\
    Qwen3-VL-32B & 50.56 & 34.26 & 63.66 & 49.00 & 50.68 & 49.63 \\
    Qwen3-Omni-30B & 49.32 & 29.37 & 54.32 & 42.68 & 41.55 & 43.45 \\
    InternVL3.5-38B & 48.00 & 28.64 & 44.05 & 34.94 & 35.16 & 38.16 \\
    \midrule
    \method{} (Ours) & \textbf{60.78} & \textbf{49.91} & \textbf{84.39} & \underline{60.21} & \textbf{62.59} & \textbf{63.58} \\
    \bottomrule
  \end{tabular*}
\end{table}

\runin{Judge calibration.}%
To compare automated scores with human judgments, we sample 100 open-text
fields, stratified by model, field type, and judge score, and ask a blinded
human rater to score them from 0 to 4. We multiply the judge scores by four
and round them to the same scale before comparing the two sets of ratings.
Fig.~\ref{fig:app_judge_trend} shows their relationship, and the table below
reports agreement statistics with bootstrap 95\% confidence intervals.
\begin{center}
\footnotesize
\renewcommand{\arraystretch}{1.12}
\begin{tabularx}{\columnwidth}{@{}Xrr@{}}
\toprule
\textbf{Statistic} & \textbf{Estimate} & \textbf{95\% CI}\\
\midrule
Spearman's $\rho$ & 0.65 & [0.49, 0.78]\\
Quadratic weighted $\kappa$ & 0.55 & [0.41, 0.67]\\
Agreement within one point & 84\% & [76\%, 91\%]\\
Mean human minus judge score & 0.65 & [0.45, 0.83]\\
\bottomrule
\end{tabularx}
\end{center}
The positive mean difference indicates that the judge assigns lower scores
than the human rater.

\providecommand{\scoringhead}[1]{\par\addvspace{3pt}\noindent\textbf{#1}\hspace{0.4em}}
\section{Scoring Protocol}
\label{app:scoring}
This appendix supplements the evaluation metrics in
Section~4.3 of the main paper, detailing field scores, task aggregation,
and the treatment of unparsed outputs.

\subsection{Field-Level Scoring}
\label{app:judge_config}
For an evaluated field $f$, let $\hat y_f$ and $y_f$ denote the prediction
and reference. The field score is
\begin{equation}
\label{eq:app_field_score}
s_f=\begin{cases}
\operatorname{F1}_{\mathrm{set}}(\hat y_f,y_f), & f\in\mathcal F_{\mathrm{cat}},\\
\mathbf{1}[\hat y_f=y_f], & f\in\mathcal F_{\mathrm{label}},\\
\bar d_{1,f}(1-\tfrac12\bar d_{2,f}), & f\in\mathcal F_{\mathrm{text}}.
\end{cases}
\end{equation}
Here $\mathcal F_{\mathrm{cat}}$ contains Affective State category fields.
$\mathcal F_{\mathrm{label}}$ contains polarity, intensity, regulation tactic,
and regulation source and target categories. $\mathcal F_{\mathrm{text}}$
contains open-text fields. Label comparisons normalize case and separators.
The indicator $\mathbf{1}[\cdot]$ is 1 for a match and 0 otherwise.
DeepSeek-V4-Flash rates each open-text field three times for T2 and twice
for the other tasks. $d_1$ and $d_2$ are averaged separately before computing the
field score.

\scoringhead{Semantic adequacy ($d_1$).}%
The judge assesses agreement with the reference in meaning, including
the character, focal episode, and information requested by the field.
Equivalent wording receives the same credit.

\begin{table}[!htbp]
\centering
\caption{Semantic adequacy $d_1$: higher is better.}
\label{tab:app_adequacy}
\small
\renewcommand{\arraystretch}{1.16}
\begin{tabularx}{\columnwidth}{@{}rX@{}}
\toprule
\textbf{Score} & \textbf{Criterion}\\
\midrule
0.00 & No meaningful match, an opposite claim, or the wrong character or episode.\\
0.25 & Only a broad or generic relation to the reference.\\
0.50 & Partial match, with an important detail or role missing or incorrect.\\
0.75 & Correct central meaning, with a secondary omission or harmless addition.\\
1.00 & Correct content, characters, and episode, regardless of wording.\\
\bottomrule
\end{tabularx}
\end{table}

\scoringhead{Redundancy ($d_2$).}%
The judge estimates how much of the answer adds no information relevant to
the queried field. This includes repetition within the answer, irrelevant
restatements of the question, generic filler, and competing guesses.
Agreement with the reference and useful supporting details are not
redundancy. Answer length itself is not penalized, and factual errors are
assessed through $d_1$.

\begin{table}[!htbp]
\centering
\caption{Redundancy $d_2$: higher is worse.}
\label{tab:app_redundancy}
\small
\renewcommand{\arraystretch}{1.16}
\begin{tabularx}{\columnwidth}{@{}rX@{}}
\toprule
\textbf{Score} & \textbf{Criterion}\\
\midrule
0.00 & Focused answer with almost no unnecessary content.\\
0.25 & Minor repetition, filler, or unnecessary hedging.\\
0.50 & About half the answer is uninformative or lists competing guesses.\\
0.75 & Most of the answer is repetition, generic content, or competing guesses.\\
1.00 & Almost entirely redundant or noncommittal.\\
\bottomrule
\end{tabularx}
\end{table}
The judge assigns each rating on a continuous $[0,1]$ scale rather than
calculating it from word counts. The bars in \eqref{eq:app_field_score}
denote the mean of the repeated ratings for each dimension.

\subsection{Task-Level Aggregation}
\label{app:scoring_details}
Field scores are combined according to the task's requested outputs,
including Regulation Evidence in Task~2.
In Task~5, the given affective object is excluded from the State score.
The equations below instantiate the hierarchy in
(6) of the main paper when all components are annotated:
\begin{align}
S_{\mathrm{T1}}&=\tfrac12(S_{\mathrm{State}}+S_{\mathrm{Manifestation}}),\notag\\
S_{\mathrm{T2}}&=\tfrac14(S_{\mathrm{Tactic}}+S_{\mathrm{Target}}+S_{\mathrm{Goal}}+S_{\mathrm{Evidence}}),\notag\\
S_{\mathrm{T3}}&=\tfrac12(S_{\mathrm{External\,Reality}}+S_{\mathrm{Internal\,Driver}}),\notag\\
S_{\mathrm{T4}}&=S_{\mathrm{requested\,Effect}},\notag\\
S_{\mathrm{T5}}&=\tfrac13(S_{\mathrm{Condition}}+S_{\mathrm{Affect}}+S_{\mathrm{Effect}}),\notag\\
S_{\mathrm{Affect}}&=\tfrac13(S_{\mathrm{State}}+S_{\mathrm{Manifestation}}+S_{\mathrm{Regulation}}).
\label{eq:score_full_chain}
\end{align}

\scoringhead{Applicable fields.}%
The reference determines which components enter each mean. Two cases must
be distinguished:
\begin{itemize}
\item \textbf{Not annotated in the reference.} Exclude the component and
renormalize the remaining weights at that level.
\item \textbf{Missing from the prediction.} Keep the component's weight
and assign it zero.
\end{itemize}
Item scores are averaged within each task, and the five task means
contribute equally to the overall score, as in
(6) of the main paper.

\scoringhead{T2 regulation.}%
The source state is supplied in the question, so the task evaluates the
regulation rather than recognizing that state again:
\begin{itemize}
\item \textbf{Target.} Score it only when the reference target category
differs from the supplied source category, that is, for Substitution and
Fabrication. Two predicted categories do
not match a single reference category.
\item \textbf{No regulation.} Score tactic alone.
\end{itemize}
Here $S_{\mathrm{Evidence}}$ scores Regulation Evidence
($M_{\mathrm{reg}}$), the manifestation cues supporting the regulation
judgment. It is a task-specific output, not a separate Blueprint construct.
For regulated items, it receives equal weight with the other applicable
fields. Neither the evidence score nor the goal score enters the mean for
items without regulation.

\subsection{Parsing and Score Reporting}
\label{app:score_reporting}
An output that cannot be parsed receives an item score of zero.
Table~\ref{tab:parse_validity} distinguishes the resulting all-item mean
from the mean over parsable outputs alone.

All reported scores keep parsing failures at zero. The table gives both
values for settings below 90\% parsing success, with the reported value
in bold.
\begin{table}[!htbp]
  \centering
  \caption{Output parsing rates below 90\%. Bold values are reported in Table~2 of the main paper.}
  \label{tab:parse_validity}
  \footnotesize
  \setlength{\tabcolsep}{2.5pt}
  \renewcommand{\arraystretch}{1.08}
  \begin{tabular*}{\linewidth}{@{\extracolsep{\fill}}llrrr@{}}
    \toprule
    \textbf{Model} & \textbf{Task} & \textbf{Pars. (\%)} & \textbf{All} & \textbf{Pars. only} \\
    \midrule
    Emotion-LLaMA\textsuperscript{*} & T2 & \expnum{61} & \textbf{\expnum{5.22}} & \expnum{8.61} \\
    Emotion-LLaMA\textsuperscript{\dag} & T3 & \expnum{37} & \textbf{\expnum{2.26}} & \expnum{6.11} \\
    Emotion-LLaMA\textsuperscript{\dag} & T4 & \expnum{49.5} & \textbf{\expnum{0.31}} & \expnum{0.63} \\
    Emotion-LLaMA\textsuperscript{\dag} & T5 & \expnum{7} & \textbf{\expnum{0.20}} & \expnum{2.81} \\
    AffectGPT\textsuperscript{\dag} & T2 & \expnum{17} & \textbf{\expnum{4.64}} & \expnum{27.38} \\
    AffectGPT\textsuperscript{*} & T3 & \expnum{79} & \textbf{\expnum{17.99}} & \expnum{22.84} \\
    AffectGPT\textsuperscript{*} & T4 & \expnum{83} & \textbf{\expnum{15.79}} & \expnum{18.94} \\
    AffectGPT\textsuperscript{*} & T5 & \expnum{87} & \textbf{\expnum{20.87}} & \expnum{23.96} \\
    Emotion-Qwen\textsuperscript{*} & T1 & \expnum{81} & \textbf{\expnum{29.45}} & \expnum{36.58} \\
    Emotion-Qwen\textsuperscript{*} & T2 & \expnum{81} & \textbf{\expnum{24.24}} & \expnum{30.09} \\
    Emotion-Qwen\textsuperscript{*} & T3 & \expnum{82} & \textbf{\expnum{21.04}} & \expnum{25.74} \\
    Emotion-Qwen\textsuperscript{*} & T5 & \expnum{77} & \textbf{\expnum{19.13}} & \expnum{24.80} \\
    LLaVA-OneVision-7B\textsuperscript{\dag} & T5 & \expnum{29} & \textbf{\expnum{5.47}} & \expnum{18.76} \\
    LLaVA-NeXT-Video-32B\textsuperscript{*} & T3 & \expnum{84} & \textbf{\expnum{27.13}} & \expnum{32.19} \\
    LLaVA-NeXT-Video-32B\textsuperscript{*} & T5 & \expnum{54} & \textbf{\expnum{13.80}} & \expnum{25.77} \\
    \bottomrule
  \end{tabular*}
  \tabnote{Pars.: share of parsable outputs. All: unparsable outputs score zero. Pars. only: mean over the parsable outputs. \textsuperscript{*}At least half parsable. \textsuperscript{\dag}Fewer than half parsable.}
\end{table}

\section{Data Construction and Annotation}
\label{app:data_quality}

\subsection{Annotation and Human Revision}
\label{app:revision_stats}

\runin{Paired records.}%
To quantify human edits, we pair 1{,}125 retained MLLM-proposed chains with
their raw human records from the annotation archive, before AI polishing.
For each field, we restrict the comparison to chains where that field
is present in both records.

\begin{figure}[!t]
  \centering
  \includegraphics[width=\columnwidth]{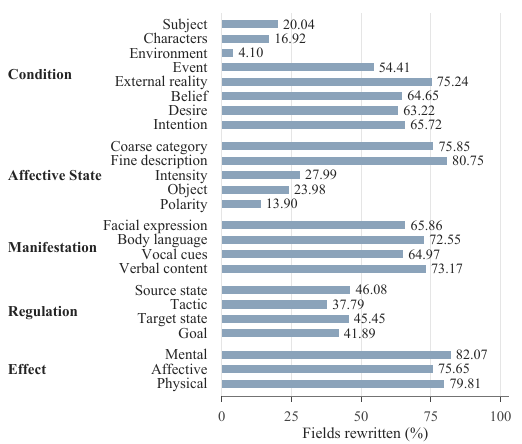}
  \caption{Field-level rewrite rates after human review of retained MLLM-proposed chains.}
  \label{fig:app_revision_flow}
\end{figure}

\runin{Edit criterion.}%
We count a field as edited when its value differs after ignoring formatting
and standardizing empty-value markers. The resulting rates in
Fig.~\ref{fig:app_revision_flow} therefore include wording changes as well
as changes in meaning.

\begin{figure*}[!t]
  \centering
  \includegraphics[width=0.90\textwidth]{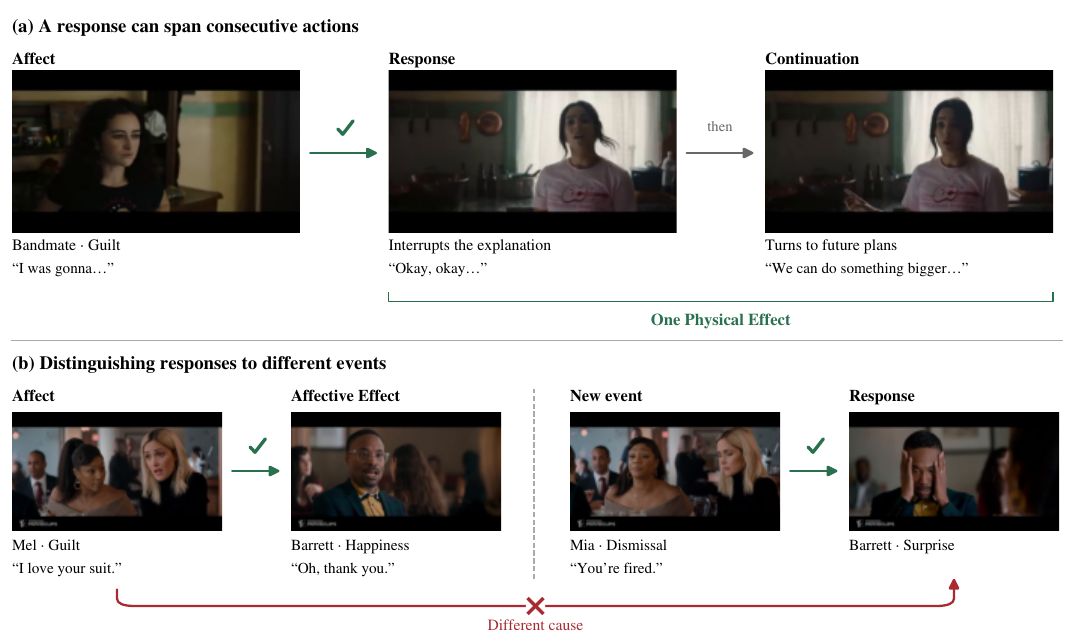}
  \caption{Effect boundaries. (a) Consecutive actions belong to one Physical Effect. (b) Barrett's later surprise follows the dismissal, not Mel's preceding Affect. Checks mark supported links, and the cross marks the incorrect attribution.}
  \label{fig:app_effect_annotation}
\end{figure*}

\begin{figure*}[!t]
  \centering
  \includegraphics[width=\textwidth]{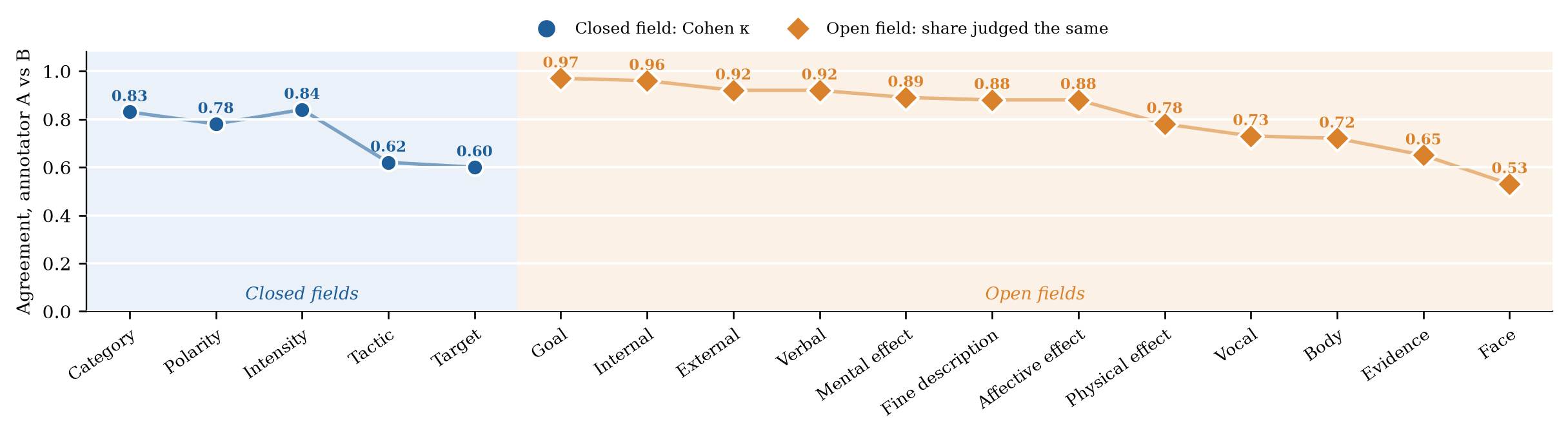}
  \caption{Agreement between two independent re-annotations. Closed fields:
  Cohen's $\kappa$ (linear weights for Intensity). Open fields: proportion
  of pairs judged to state the same content. Only $\kappa$ corrects for
  chance agreement; the two measures are not directly comparable.}
  \label{fig:app_iaa_ab}
\end{figure*}

\subsection{Annotation Consistency}
\label{app:iaa}

Dataset construction involved eight trained annotators.
Two annotators independently re-annotated 50 chains excluded from guideline
development, using frames and subtitles. We compare these two annotations
(A and B) field by field, including only chains for which both provide an
applicable answer. Fig.~\ref{fig:app_iaa_ab} reports the results.

\runin{Closed fields.}%
We compute Cohen's $\kappa$, with linear weights for the ordered intensity
levels (Low, Medium, and High). Category, Tactic, and Intensity each have
50 eligible chains; Polarity has 32 and Target has 29. For Category, we
compare the first listed emotion category. For Target, we compare only
the emotion category, taking it to match the felt category for Suppression
and Amplification.

\runin{Open-text fields.}%
For each field, we pair the non-empty texts from A and B and randomize
their order. A model judge (\texttt{claude-opus-5}), blind to annotator
identity, assesses the pairs using a shared rubric: the same
content (1), partly overlapping (0.5), or different or contradictory
(0). Wording and length may differ without changing the content; adding or
omitting a substantive element constitutes partial agreement. We report
strict agreement, the proportion of eligible pairs rated 1, rather than
the mean rating. This measures agreement on content where both annotations
provide text, not on whether a field should be present.

Eligible pair counts are 50 each for External Reality, Internal Driver,
fine-grained affective description, and Body Language; 49 each for Facial
Expression and Vocal Cues; 48 for Verbal Content; 31 each for Regulation
Goal and Evidence; and 45, 33, and 46 for Mental, Affective, and Physical
Effect, respectively. Counts fall below 50 when a chain lacks the field
or either annotator leaves it unfilled.

\needspace{5\baselineskip}
\subsection{Field Annotation Guidance}
\label{app:field_guidance}
The following criteria supplement the annotation procedure in the main paper.

\par\addvspace{4pt}
\noindent\textbf{Intensity.}\hspace{0.4em}%
State intensity is judged from the situation, the person's concerns, and
their response. A calm Manifestation or successful Regulation does not imply
low intensity, and high intensity does not require loss of control. The
levels are:
\begin{itemize}[leftmargin=1.3em,itemsep=1pt,parsep=0pt,topsep=3pt]
\item \textbf{Low}: mild feeling with little influence on attention or ongoing activity.
\item \textbf{Medium}: feeling that changes attention or response without dominating the episode.
\item \textbf{High}: strong feeling that dominates the person's concerns or creates an urgent need to respond.
\end{itemize}
\par\vspace{3pt}

\runin{Mixed polarity.}%
Positive and negative feelings coexist in the same person at the same moment,
such as relief with embarrassment. A mismatch between State and Manifestation,
or a change across moments, is not sufficient.

\runin{Effect.}%
When annotating an Effect, we include connected utterances or actions that
respond to the subject's affective episode, even when they extend beyond
the first response. We exclude reactions to a new event, as illustrated
by the contrasting cases in Fig.~\ref{fig:app_effect_annotation}.

\subsection{Task and Answer Formats}
\label{app:task_questions}
Following Section~4.1 of the main paper, each question combines episode context
($\mathrm{ctx}_t$), known constructs ($K_t$), and a JSON template for queried
constructs ($Q_t$). Task~2 stores Regulation Evidence ($M_{\mathrm{reg}}$)
in the \texttt{evidence} field of \texttt{regulation\_pathway}; this is not
an additional Blueprint construct. Fig.~\ref{fig:app_task_templates}
gives the question and answer formats for all five tasks.

\needspace{7\baselineskip}
\begingroup
\captionof{figure}{Task and answer templates. Braces indicate item-specific content.
Field names follow Table~1 of the main paper. Item identifiers
are omitted. Manifestation channels, Regulation, Internal Driver, and
Effect channels appear only when annotated.}
\label{fig:app_task_templates}
\begin{taskframe}{T1  Grounded Affect Recognition}
\plabel{Task}Infer the subject's affective state and observable affective manifestation.\par\pgap
\plabel{Focal Context}Focus on \{subject\}. \{scene and trigger\}, based on the subject's observed \{annotated channels\}, infer the subject's internal affective state.\par\pgap
\plabel{Known Structured Fields}\par{\raggedright\hangindent=1em\hangafter=1 \texttt{- subject: \{subject\}}\par}\pgap
\plabel{Answer Template}\par{\raggedright
\texttt{\{}\\
\hspace*{1.0em}\texttt{{\textquotedbl}subject{\textquotedbl}: {\textquotedbl}\{subject\}{\textquotedbl},}\\
\hspace*{1.0em}\texttt{{\textquotedbl}affective\_state{\textquotedbl}: \{}\\
\hspace*{2.0em}\texttt{{\textquotedbl}object{\textquotedbl}: {\textquotedbl}string{\textquotedbl}, {\textquotedbl}coarse\_category{\textquotedbl}: [{\textquotedbl}string{\textquotedbl}],}\\
\hspace*{2.0em}\texttt{{\textquotedbl}fine\_description{\textquotedbl}: {\textquotedbl}string{\textquotedbl},}\\
\hspace*{2.0em}\texttt{{\textquotedbl}polarity{\textquotedbl}: {\textquotedbl}string{\textquotedbl}, {\textquotedbl}intensity{\textquotedbl}: {\textquotedbl}string{\textquotedbl} \},}\\
\hspace*{1.0em}\texttt{{\textquotedbl}affective\_manifestation{\textquotedbl}: \{}\\
\hspace*{2.0em}\texttt{{\textquotedbl}facial\_expression{\textquotedbl}: {\textquotedbl}string{\textquotedbl},}\\
\hspace*{2.0em}\texttt{{\textquotedbl}body\_language{\textquotedbl}: {\textquotedbl}string{\textquotedbl},}\\
\hspace*{2.0em}\texttt{{\textquotedbl}vocal\_cues{\textquotedbl}: {\textquotedbl}string{\textquotedbl}, {\textquotedbl}verbal\_content{\textquotedbl}: {\textquotedbl}string{\textquotedbl} \}}\\
\texttt{\}}\par}
\end{taskframe}
\begin{taskframe}{T2  Regulation Decoding}
\plabel{Task}Given the subject's affective state (category and intensity) at the focal moment, decide whether and how the subject regulates its outward manifestation. Report the regulation tactic, the goal, the target affective state as a category, and the evidence observed in the video.\par\pgap
\plabel{Focal Context}Focus on \{subject\}. Focal window: \{m:ss\}-\{m:ss\}.\par\pgap
\plabel{Known Structured Fields}\par{\raggedright\hangindent=1em\hangafter=1 \texttt{- source\_subject: \{subject\}}\par\hangindent=1em\hangafter=1 \texttt{- affective\_state.object: \{object\}}\par\hangindent=1em\hangafter=1 \texttt{- affective\_state.coarse\_category: \{category\}}\par\hangindent=1em\hangafter=1 \texttt{- affective\_state.intensity: \{intensity\}}\par}\pgap
\plabel{Answer Template}\par{\raggedright
\texttt{\{}\\
\hspace*{1.0em}\texttt{{\textquotedbl}subject{\textquotedbl}: {\textquotedbl}\{subject\}{\textquotedbl},}\\
\hspace*{1.0em}\texttt{{\textquotedbl}regulation\_pathway{\textquotedbl}: \{}\\
\hspace*{2.0em}\texttt{{\textquotedbl}tactic\_type{\textquotedbl}: {\textquotedbl}string{\textquotedbl}, {\textquotedbl}goal{\textquotedbl}: {\textquotedbl}string{\textquotedbl},}\\
\hspace*{2.0em}\texttt{{\textquotedbl}target\_affect\_state{\textquotedbl}: {\textquotedbl}string{\textquotedbl}, {\textquotedbl}evidence{\textquotedbl}: {\textquotedbl}string{\textquotedbl} \}}\\
\texttt{\}}\par}
\end{taskframe}
\begin{taskframe}{T3  Affective Cause Reasoning}
\plabel{Task}Given the subject's affective state and manifestation, infer the affective cause requested by the answer template.\par\pgap
\plabel{Focal Context}Analyze the affective cause for \{subject\}'s \{affect\} toward \{object\}.\par\pgap
\plabel{Known Structured Fields}\par{\raggedright\hangindent=1em\hangafter=1 \texttt{- source\_subject: \{subject\}}\par\hangindent=1em\hangafter=1 \texttt{- affective\_state.object: \{object\}}\par\hangindent=1em\hangafter=1 \texttt{- affective\_state.coarse\_category: \{category\}}\par\hangindent=1em\hangafter=1 \texttt{- affective\_state.polarity: \{polarity\}}\par\hangindent=1em\hangafter=1 \texttt{- affective\_state.intensity: \{intensity\}}\par\hangindent=1em\hangafter=1 \texttt{- affective\_manifestation.facial\_expression: \{text\}}\par\hangindent=1em\hangafter=1 \texttt{- affective\_manifestation.body\_language: \{text\}}\par\hangindent=1em\hangafter=1 \texttt{- affective\_manifestation.vocal\_cues: \{text\}}\par\hangindent=1em\hangafter=1 \texttt{- affective\_manifestation.verbal\_content: \{text\}}\par\hangindent=1em\hangafter=1 \texttt{- regulation\_pathway.tactic\_type: \{tactic\}}\par\hangindent=1em\hangafter=1 \texttt{- regulation\_pathway.goal: \{text\}}\par\hangindent=1em\hangafter=1 \texttt{- regulation\_pathway.target\_affect\_state: \{category\}}\par\hangindent=1em\hangafter=1 \texttt{- regulation\_pathway.evidence: \{text\}}\par}\pgap
\begin{minipage}{\linewidth}
\plabel{Answer Template}\par{\raggedright
\texttt{\{}\\
\hspace*{1.0em}\texttt{{\textquotedbl}subject{\textquotedbl}: {\textquotedbl}\{subject\}{\textquotedbl},}\\
\hspace*{1.0em}\texttt{{\textquotedbl}affective\_cause{\textquotedbl}: \{}\\
\hspace*{2.0em}\texttt{{\textquotedbl}internal\_driver{\textquotedbl}: {\textquotedbl}string{\textquotedbl},}\\
\hspace*{2.0em}\texttt{{\textquotedbl}external\_reality{\textquotedbl}: {\textquotedbl}string{\textquotedbl} \}}\\
\texttt{\}}\par}
\end{minipage}
\end{taskframe}
\begin{taskframe}{T4  Affective Effect Reasoning}
\plabel{Task}Given the source full affect stage, infer the requested downstream effect on the specified recipient.\par\pgap
\plabel{Focal Context}Known source affective episode: \{subject\}'s full affect toward \{object\}. Requested downstream effect: \{effect type\} on \{recipient\}.\par\pgap
\plabel{Known Structured Fields}\par{\raggedright\hangindent=1em\hangafter=1 \texttt{- source\_subject: \{subject\}}\par\hangindent=1em\hangafter=1 \texttt{- source\_affective\_state.object: \{object\}}\par\hangindent=1em\hangafter=1 \texttt{- source\_affective\_state.coarse\_category: \{category\}}\par\hangindent=1em\hangafter=1 \texttt{- source\_affective\_state.fine\_description: \{text\}}\par\hangindent=1em\hangafter=1 \texttt{- source\_affective\_state.polarity: \{polarity\}}\par\hangindent=1em\hangafter=1 \texttt{- source\_affective\_state.intensity: \{intensity\}}\par\hangindent=1em\hangafter=1 \texttt{- source\_affective\_manifestation.facial\_expression: \{text\}}\par\hangindent=1em\hangafter=1 \texttt{- source\_affective\_manifestation.body\_language: \{text\}}\par\hangindent=1em\hangafter=1 \texttt{- source\_affective\_manifestation.vocal\_cues: \{text\}}\par\hangindent=1em\hangafter=1 \texttt{- source\_affective\_manifestation.verbal\_content: \{text\}}\par\hangindent=1em\hangafter=1 \texttt{- source\_regulation\_pathway.tactic\_type: \{tactic\}}\par\hangindent=1em\hangafter=1 \texttt{- source\_regulation\_pathway.goal: \{text\}}\par\hangindent=1em\hangafter=1 \texttt{- source\_regulation\_pathway.target\_affect\_state: \{category\}}\par\hangindent=1em\hangafter=1 \texttt{- source\_regulation\_pathway.evidence: \{text\}}\par\hangindent=1em\hangafter=1 \texttt{- effect\_type: \{effect type\}}\par\hangindent=1em\hangafter=1 \texttt{- effect\_recipient: \{recipient\}}\par}\pgap
\plabel{Answer Template}\par{\raggedright
\texttt{\{}\\
\hspace*{1.0em}\texttt{{\textquotedbl}subject{\textquotedbl}: {\textquotedbl}\{subject\}{\textquotedbl},}\\
\hspace*{1.0em}\texttt{{\textquotedbl}object{\textquotedbl}: {\textquotedbl}\{object\}{\textquotedbl},}\\
\hspace*{1.0em}\texttt{{\textquotedbl}\{effect type\}{\textquotedbl}: {\textquotedbl}string{\textquotedbl}}\\
\texttt{\}}\par}
\end{taskframe}
\begin{taskframe}{T5  Full Chain Reconstruction}
\plabel{Task}Reconstruct the complete condition-affect-effect chain for the focal affective moment.\par\pgap
\plabel{Focal Context}Reconstruct the affective chain for \{subject\}'s affect related to \{object\} in the environment \{environment\} Around this focal moment: When \{name\} says ``\{line\}''.\par\pgap
\plabel{Known Structured Fields}\par{\raggedright\hangindent=1em\hangafter=1 \texttt{- subject: \{subject\}}\par\hangindent=1em\hangafter=1 \texttt{- affect\_object: \{object\}}\par\hangindent=1em\hangafter=1 \texttt{- focal\_anchor: Around this focal moment: When \{name\} says {\textquotedbl}\{line\}{\textquotedbl}.}\par\hangindent=1em\hangafter=1 \texttt{- given\_affective\_fields: none; reconstruct the full chain from the focal anchor and multimodal evidence}\par}\pgap
\plabel{Answer Template}\par{\raggedright
\texttt{\{}\\
\hspace*{1.0em}\texttt{{\textquotedbl}1\_condition{\textquotedbl}: \{}\\
\hspace*{2.0em}\texttt{{\textquotedbl}subject{\textquotedbl}: {\textquotedbl}\{subject\}{\textquotedbl},}\\
\hspace*{2.0em}\texttt{{\textquotedbl}causal\_analysis{\textquotedbl}: \{}\\
\hspace*{3.0em}\texttt{{\textquotedbl}internal\_driver{\textquotedbl}: {\textquotedbl}<TBI>{\textquotedbl},}\\
\hspace*{3.0em}\texttt{{\textquotedbl}external\_reality{\textquotedbl}: {\textquotedbl}<TBI>{\textquotedbl} \} \},}\\
\hspace*{1.0em}\texttt{{\textquotedbl}2\_affect{\textquotedbl}: \{}\\
\hspace*{2.0em}\texttt{{\textquotedbl}affective\_state{\textquotedbl}: \{}\\
\hspace*{3.0em}\texttt{{\textquotedbl}object{\textquotedbl}: {\textquotedbl}\{object\}{\textquotedbl}, {\textquotedbl}coarse\_category{\textquotedbl}: {\textquotedbl}<TBI>{\textquotedbl},}\\
\hspace*{3.0em}\texttt{{\textquotedbl}fine\_description{\textquotedbl}: {\textquotedbl}<TBI>{\textquotedbl},}\\
\hspace*{3.0em}\texttt{{\textquotedbl}intensity{\textquotedbl}: {\textquotedbl}<TBI>{\textquotedbl}, {\textquotedbl}polarity{\textquotedbl}: {\textquotedbl}<TBI>{\textquotedbl} \},}\\
\hspace*{2.0em}\texttt{{\textquotedbl}affective\_manifestation{\textquotedbl}: \{}\\
\hspace*{3.0em}\texttt{{\textquotedbl}facial\_expression{\textquotedbl}: {\textquotedbl}<TBI>{\textquotedbl},}\\
\hspace*{3.0em}\texttt{{\textquotedbl}body\_language{\textquotedbl}: {\textquotedbl}<TBI>{\textquotedbl},}\\
\hspace*{3.0em}\texttt{{\textquotedbl}vocal\_cues{\textquotedbl}: {\textquotedbl}<TBI>{\textquotedbl}, {\textquotedbl}verbal\_content{\textquotedbl}: {\textquotedbl}<TBI>{\textquotedbl} \},}\\
\hspace*{2.0em}\texttt{{\textquotedbl}regulation\_pathway{\textquotedbl}: \{}\\
\hspace*{3.0em}\texttt{{\textquotedbl}tactic\_type{\textquotedbl}: {\textquotedbl}<TBI>{\textquotedbl}, {\textquotedbl}goal{\textquotedbl}: {\textquotedbl}<TBI>{\textquotedbl},}\\
\hspace*{3.0em}\texttt{{\textquotedbl}source\_affect\_state{\textquotedbl}: {\textquotedbl}<TBI>{\textquotedbl},}\\
\hspace*{3.0em}\texttt{{\textquotedbl}target\_affect\_state{\textquotedbl}: {\textquotedbl}<TBI>{\textquotedbl} \} \},}\\
\hspace*{1.0em}\texttt{{\textquotedbl}3\_effect{\textquotedbl}: \{}\\
\hspace*{2.0em}\texttt{{\textquotedbl}mental\_effect{\textquotedbl}: {\textquotedbl}<TBI> on \{recipient\}{\textquotedbl},}\\
\hspace*{2.0em}\texttt{{\textquotedbl}affective\_effect{\textquotedbl}: {\textquotedbl}<TBI> on \{recipient\}{\textquotedbl},}\\
\hspace*{2.0em}\texttt{{\textquotedbl}physical\_effect{\textquotedbl}: {\textquotedbl}<TBI> on \{recipient\}{\textquotedbl} \}}\\
\texttt{\}}\par}
\end{taskframe}

\endgroup